\documentclass[letterpaper, 10 pt, conference]{ieeeconf}  

\IEEEoverridecommandlockouts                              

\usepackage{graphics} 
\usepackage{epsfig} 
\usepackage{newtxmath} 
\usepackage{times} 
\usepackage{cite}
\usepackage{amsmath}

\usepackage{amssymb}
\usepackage{algorithmic}
\usepackage{subfigure}
\usepackage{textcomp}
\usepackage{xcolor}
\usepackage{booktabs}
\usepackage{multirow}
\usepackage{graphicx}

\title{\LARGE \bf
Direction and Trajectory Tracking Control for Nonholonomic Spherical Robot by Combining Sliding Mode Controller and Model Prediction Controller*
}

\author{Yifan Liu$^{1}$, Yixu Wang$^{1}$, Xiaoqing Guan$^{1}$, Tao Hu$^{1}$, Ziang Zhang$^{1}$, Song Jin$^{1}$,\\You Wang$^{1, \dagger}$, Jie Hao$^{2}$ and Guang Li$^{1}$
\thanks{*This work was supported by the 
National Key Research and Development Program of China 
(No. SQ2019YFB130016) and 
the Open Research Project of the State Key Laboratory of 
Industrial Control Technology (No. ICT2021B10).}
\thanks{$^{1}$
State Key Laboratory of Industrial Control Technology, Institute of Cyber Systems and Control, 
Zhejiang University, Hangzhou, 310027, China.}
\thanks{$^{2}$Jie Hao is with Luoteng Hangzhou Techonlogy Co.,Ltd., Hangzhou, 310027, China.}
\thanks{$\dagger$ Corresponding author is You Wang, e-mail: king\_wy@zju.edu.cn.
}
}

\begin{document}

\maketitle
\thispagestyle{empty}
\pagestyle{empty}

\begin{abstract}
Spherical robot is a nonlinear, nonholonomic and unstable system which increases the difficulty of the direction and trajectory tracking problem. In this study, we propose a new direction controller HTSMC, an instruction planning controller MPC, and a trajectory tracking framework MHH. The HTSMC is designed by integrating a fast terminal algorithm, a hierarchical method, the motion features of a spherical robot, and its dynamics. In addition, the new direction controller has an excellent control effect with a quick response speed and strong stability. MPC can obtain optimal commands  that are then transmitted to the velocity and direction controller. Since the two torque controllers in MHH are all Lyapunov-based sliding mode controllers, the MHH framework may achieve optimal control performance while assuring stability. Finally, the two controllers eliminate the requirement for MPC's stability and dynamic constraints. Finally, hardware experiments demonstrate the efficacy of the HTSMC, MPC, and MHH.

\end{abstract}

\section{INTRODUCTION}

The spherical robot is a novel kind of mobile robot which can be used in a range of fields such as environmental detection, search and rescue, security patrols and entertainment. Compared with other mobile robots like wheeled robots and legged robots, spherical robots have several advantages \cite{bib1, bib3, bib5, bib10}. For example, because of its spherical rolling mechanism, the spherical robot can have flexible omnidirectional mobility and minimal frictional energy consumption. Furthermore, its closed spherical shell can aid in protecting the inside electrical system and mechanical framework from collision and damage \cite{bib14}. However, despite its advantages, the spherical robot is difficult to control due to its non-holonomic, non-linear and under-actuated characteristics\cite{bib17}.

\begin{figure}[tb]
\centering
\includegraphics[width=8cm]{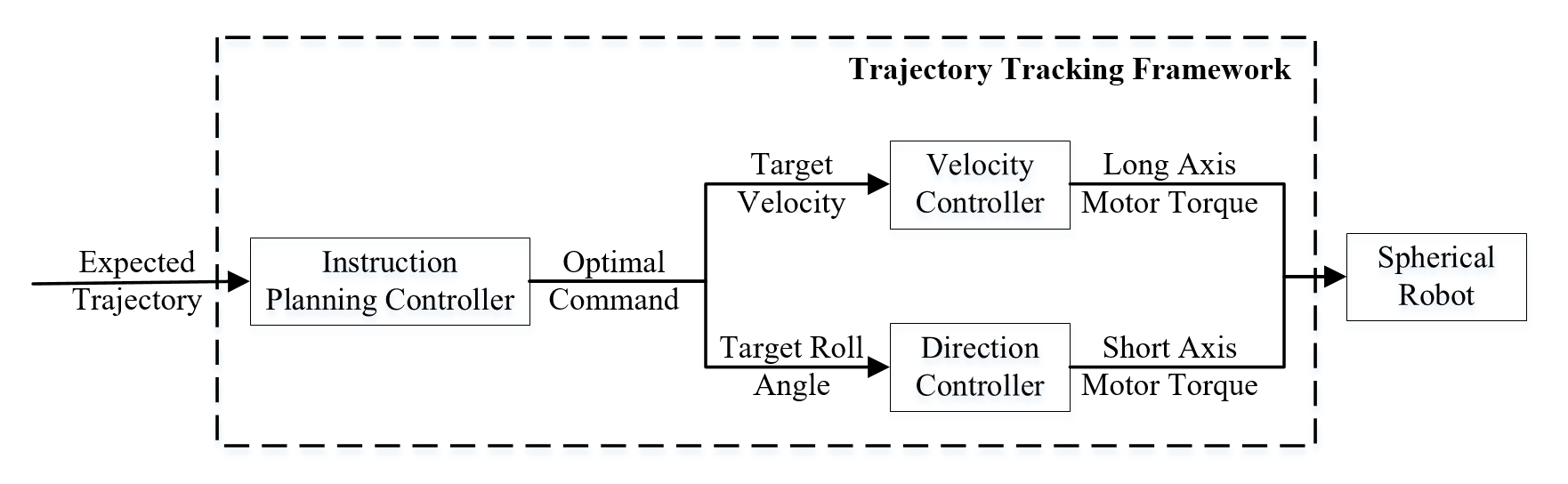}
\caption{Trajectory Tracking Framework of the spherical robot.}
\label{fig1}
\vspace{-0.5cm}
\end{figure}

Trajectory tracking control plays an important role in the robot's motion and planning. A comprehensive trajectory tracking control framework includes three portions, i.e., velocity controller and direction controller which have a close relationship with dynamic models, and instruction planning controller which transmits instructions to torque controllers mentioned above. Fig. \ref{fig1} illustrates the framework of the tracking mechanism. An efficient velocity controller have been presented in our previous work in \cite{pre1, pre2}. In this study, we focus primarily on constructing the direction controller and instruction planning controller to realize trajectory tracking control of the robot. In addition, the spherical robot's direction control system is a complex, delayed, non-holonomic, and nonlinear system that demands the design of an efficient controller. During trajectory tracking control, stability is extremely challenging, yet crucial.

Over the last few years, a series of works have been made on the control of the spherical robot. The majority of study concentrates on the spherical robot's velocity control \cite{bib3, bib4, bib22, bib23, bib1, bib9, bib10}.Several studies, for instance, develop a velocity controller based on feedback control, such as a PI-type fuzzy logic controller \cite{bib22}. Linear quadratic regulator (LQR) and sliding mode controller (SMC) are also utilized in the design of velocity controllers \cite{bib1, bib9}. Some studies develop a velocity controller by combining integral approach, adaptive law, and SMC \cite{bib10}. Adaptive velocity controllers for spherical robots are also available by predicting the uncertainty \cite{bib3, bib4, pre2}. The path tracking of spherical robots has also been the subject of a number of studies \cite{bib14, bib16}. For instance, \cite{bib16} presents a combination of fuzzy logic and SMC algorithm. There are also research focusing on motion planning of spherical robots with optimal planners that place a strong emphasis on obstacles \cite{bib28, bib29}.
The majority of spherical robot direction controllers only have simulation results, such as the direction control component in \cite{bib14, bib16, bib19}. A full-state feedback controller for roll motion is provided in \cite{bib30}. Due to the lack of an experiment-verified dynamic model for spherical robots, PID controllers are widely used for these robots and an efficient Fuzzy-PID controller is proposed in \cite{pre3} last year. However, the nonholonomic property of the spherical robot created feedback control problems \cite{bib23}. In our prior work, a hierarchical sliding mode controller (HSMC) was devised for both velocity and direction control problems. As a complex nonlinear second-order system, however, the direction controller failed to operate, and oscillation is relatively easy to happen. Hierarchical control methods are advantageous for non-holonomic and under-actuated SIMO systems \cite{bib24}. And fast terminal sliding mode controller is a type of SMC that may increase the response speed while maintaining stability \cite{bib31, bib32}. These may assist us in resolving the preceding issue.

There are other studies on trajectory tracking and instruction planning \cite{bib33, bib19, bib34}. However, they only provide simulation results without on hardware experiments. Model prediction controller (MPC) is a kind of effective instruction planning controller for mobile robot's trajectory tracking \cite{bib35, bib36, bib37}. As it is quite easy for spherical robots to become unstable, it is crucial to obtain optimal control performance while preserving stability \cite{bib38}.

We propose a trajectory tracking control framework MPC-HSMC-HTSMC (MHH) of spherical robot, as well as a direction controller hierarchical terminal sliding mode controller (HTSMC) and instruction planning controller MPC. Two Lyapnov-based SMCs in the framework guarantee the system's stability, hence eliminating the need to add stability constraints to the instruction planning controller MPC. Specifically, this study makes the following contributions:
\begin{itemize}
    \item We designed the spherical robot's direction controller HTSMC by combining fast terminal sliding mode algorithm, hierarchical method and the spherical robot's motion features. As far as we know, this is the first model-based direction controller for spherical robot that is applied on hardware.
    
    \item An integral trajectory tracking control framework is proposed for spherical robot, which is the first time that such a problem is solved on hardware for spherical robot. Two Lyapnov-based SMCs assure system stability, which helps simplify the nonlinear program in MPC.
    
    \item An instruction planning controller MPC is presented for spherical robot. And this is the first time that such a control scheme is applied on hardware for spherical robot.
    
    \item Practical experiments are conducted to demonstrate the controllers' effectiveness. And the whole-body dynamic model of the spherical robot has been given.
\end{itemize}

The paper is organized as follows: Section \uppercase\expandafter{\romannumeral2} establishes the spherical robot's kinematic and whole-body dynamic models and discusses the delayed problem. Section \uppercase\expandafter{\romannumeral3} designs a trajectory tracking control framework of the spherical robot. In section \uppercase\expandafter{\romannumeral4}, practical experiments are carried out, results and comparisons are presented. Finally, conclusions are given in Section \uppercase\expandafter{\romannumeral5}.

\section{MATHEMATICAL MODELS}
\begin{figure}[tb]
\centering
\includegraphics[width=7cm]{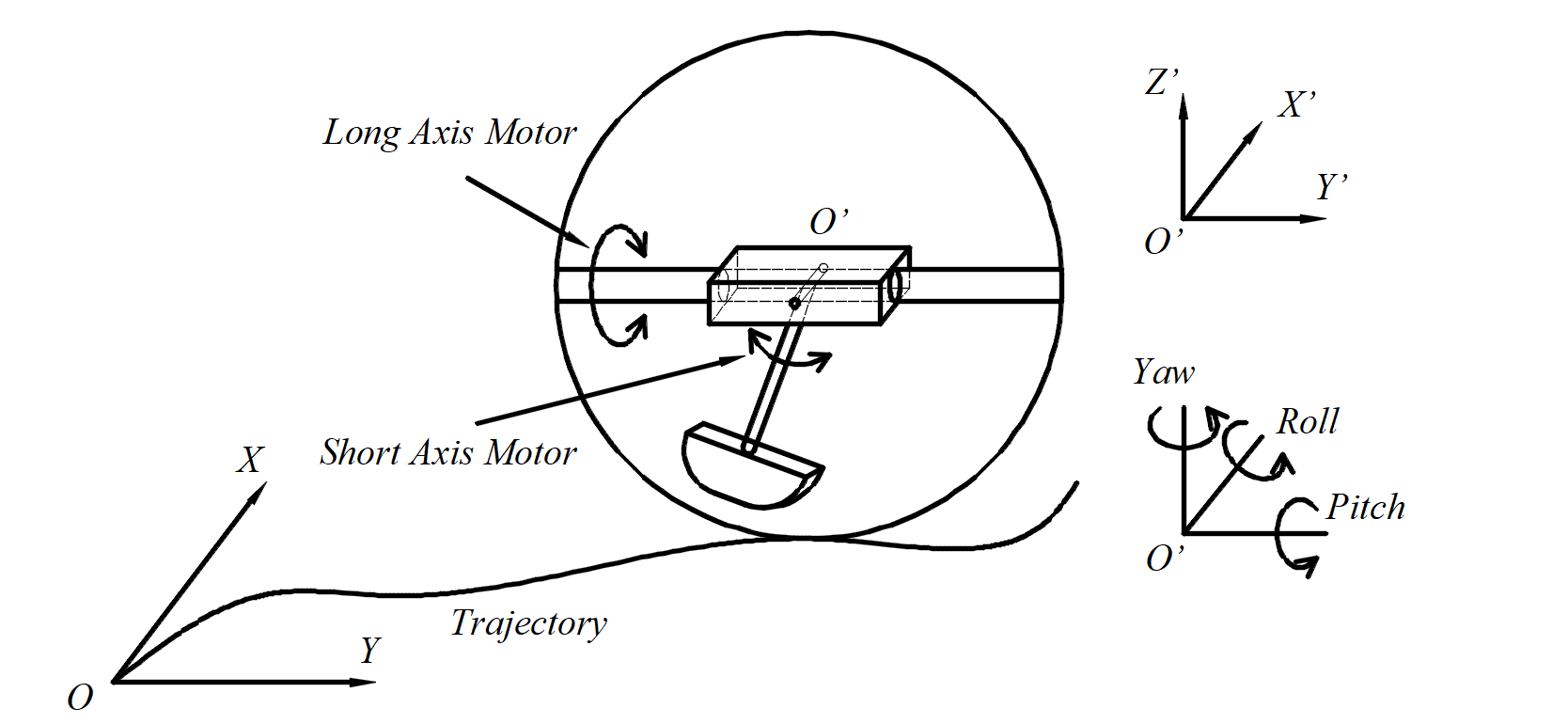}
\caption{Mechanical schematic of the spherical robot.}
\label{fig2}
\vspace{-0.5cm}
\end{figure}

The spherical robot is driven by a 2-DOF heavy pendulum and the mechanical schematic of the robot can be seen in Fig.~\ref{fig2}. A 2-DOF heavy pendulum driven by two motors is employed to drive the robot. The long axis motor is employed to control the velocity while and the short axis motor is employed to control the direction \cite{bib1}. As a result, the spherical robot may perform omnidirectional motion on a plane by adjusting the output torque of the two motors\cite{bib10}. 

To realise trajectory tracking control, firstly we need to derive the kinematic model and dynamic model. With the kinematic model and instruction planning controller, the robot can know the optimal instructions such as velocity and moving direction to track the trajectory with minimum loss. Compared with that, the dynamic model and the other two controllers decides the output torque to execute the instruction as soon as possible.

\subsection{Kinematic Model}
Assuming the robot moves on a flat terrain without slipping, according to \cite{bib28,bib29, pre3}, the kinematic model of the spherical robot can be obtained as follows:
\begin{equation}
\begin{cases}
\dot{X}=v\cos\phi-R\dot q_r \sin\phi\\
\dot{Y}=v\sin\phi+R\dot q_r \cos\phi\\
\dot{\phi}=v\tan q_r/R
\label{eq1}
\end{cases}
\end{equation}

Restricted by the central frame of the spherical robot, it cannot keep rotating in the direction of the roll angle. In this scenario, velocity $v$ is primarily responsible for determining the moving distance X and Y. Thus \eqref{eq1} can be simplified as follows:
\begin{equation}
\begin{cases}
\dot{X}=v\cos\phi\\
\dot{Y}=v\sin\phi\\
\dot{\phi}=v\tan q_r/R
\label{eq2}
\end{cases}
\end{equation}

\subsection{Whole-body Dynamic Model}
\newcommand{\tabincell}[2]{\begin{tabular}{@{}#1@{}}#2\end{tabular}} 
\linespread{1.2}
\begin{table}[tb]
\caption{List of Nomenclature}
\begin{center}
\resizebox{1.\columnwidth}{!}{
\begin{tabular}{cc}
\toprule
\textbf{Symbol}&\textbf{Description} \\
\midrule
$\phi$ & Yaw angle of the robot \\
$\alpha$, $\beta$ & Swing angle of the pendulum in pitch and roll direction\\
$\varphi$, $q_r$ & Pitch angle and roll angle of the robot \\
$x$ & \tabincell{c}{Distance that the center of the spherical robot moves,\\and $x = \varphi R$} \\
$v$ & Forward velocity of the spherical robot, and $v = \dot{x}$ \\
$\tau_v$, $\tau_r$ & Output torque of the long axis motor and short axis motor\\
$m$ & Mass of heavy pendulum \\
$m_f$ & Mass of frame \\
$m_s$ & Mass of shell \\
$I_{mp}$, $I_{mr}$& Moment of inertia of heavy pendulum in pitch and roll direction\\
$I_{fp}$, $I_{fr}$ & Moment of inertia of frame in pitch and roll direction \\
$I_sp$, $I_r$ & Moment of inertia of shell in pitch and roll direction \\
$R$ & Radius of spherical robot \\
$l$ & \tabincell{c}{Distance between the center of the spherical robot\\and the center of the pendulum} \\
$F_{fp}$ $F_{fr}$& Friction between shell and ground in $X'$ and $Y'$ direction\\
$\zeta$ & Viscous damping coefficient \\
\bottomrule
\end{tabular}
}
\label{table1}
\end{center}
\vspace{-0.5cm}
\end{table}
The schematic illustration of the spherical robot is given in Fig.~\ref{fig2}. The spherical robot can be simplified into three components: a heavy pendulum, a spherical shell, and a central frame. Variables of the robot are represented in Table.~\ref{table1}. The Euler-Lagrange method is utilized to construct the spherical robot's whole-body dynamic model. Let $L$ and $\varPsi$ represent the lagrangian function and the Rayleigh’s dissipation function of the spherical robot, the Euler-Lagrange equations can be written as follows:
\begin{equation}
\frac{\mathrm{d}}{\mathrm{d}t}\left(\frac{\partial L}{\partial \dot q_k}\right)-\frac{\partial L}{\partial q_k}+\frac{\partial \varPsi}{\partial \dot q}=\tau_{q_k}\label{eq3}
\end{equation}
where $q_k \in \boldsymbol{q}= \begin{bmatrix}\alpha&x&\beta&q_r\end{bmatrix}^T$, and $\tau_{q_k}$ is the related external torque.

Transform \eqref{eq3} into matrix form, then we can get the whole-body dynamic model of the robot as follows:
\begin{equation}
\boldsymbol{M}\left(\boldsymbol{q}\right)\boldsymbol{\ddot{q}}+\boldsymbol{N}\left(\boldsymbol{q},\boldsymbol{\dot{q}}\right)=\boldsymbol{E}\boldsymbol{\tau}\label{eq4}
\end{equation}
where $\boldsymbol{M}\left(\boldsymbol{q}\right)\in{\mathbb{R}^{4\times4}}$ is the inertia matrix, $\boldsymbol{N}\left(\boldsymbol{q},\boldsymbol{\dot{q}}\right)\in{\mathbb{R}^{4}}$ is the nonlinear matrix and $\boldsymbol{q}$ is the input matrix. And

\begin{equation}
\boldsymbol{M}\left(\boldsymbol{q}\right)=
\begin{bmatrix}
I_{mp}+I_{fp} & ml\cos{\alpha} & 0 & 0 \\
mRl\cos{\alpha} & MR+I_s/R & 0 & 0\\
0 & 0 & I_{mr} & mRl\cos\beta \\
0 & 0 & 0 & I_{sr}+I_{fr}+MR^2
\end{bmatrix} \notag
\end{equation}
\begin{equation}
\boldsymbol{N}\left(\boldsymbol{q},\boldsymbol{\dot{q}}\right)=\begin{bmatrix}
mgl\sin{\alpha}\cos{\beta}+\zeta\left(\dot{\alpha}+\dot{x}\cos{\alpha}/R\right) \\ 
-mRl\dot{\alpha}^2\sin{\alpha}+\zeta\left(\dot{\alpha}\cos{\alpha}+\dot{x}/R\right)+F_{fp}R\\
mgl\cos{\alpha}\sin{\beta}+\zeta\left(\dot{\beta}+\dot{q_r}\cos{\beta}\right)\\
-mRl\dot{\beta}^2\sin{\beta}+\zeta\left(\dot{q_r}+\dot{\beta}\cos{\beta}\right)+F_{fr}R
\end{bmatrix}\notag    
\end{equation}
\begin{equation}
\boldsymbol{E}=\begin{bmatrix}
1 & 1 & 0 & 0\\
0 & 0 & 1 & 1
\end{bmatrix}^T,
\boldsymbol{\tau}=\begin{bmatrix}
\tau_p\\\tau_r
\end{bmatrix}.\notag    
\end{equation}

The whole-body dynamic model can be decomposed into two 2D sub-models according to the motor torque, which are used for designing velocity controller and direction controller, respectively. Two sub-models are shown as follows:
\begin{equation}
\boldsymbol{M_p}\cdot\boldsymbol{\ddot{q}_p}+\boldsymbol{N_p}=\begin{bmatrix}1 & 1\end{bmatrix}^T\cdot{\tau_p}\label{eq5}
\end{equation}
\begin{equation}
\boldsymbol{M_r}\cdot\boldsymbol{\ddot{q}_r}+\boldsymbol{N_r}=\begin{bmatrix}1 & 1\end{bmatrix}^T\cdot{\tau_r}\label{eq6}
\end{equation}
where $\boldsymbol{M_p}, \boldsymbol{M_r}\in{\mathbb{R}^{2\times2}}$, and $\boldsymbol{M}=diag\begin{bmatrix}\boldsymbol{M_p}, &  \boldsymbol{M_r}\end{bmatrix}$. $\boldsymbol{N_p}, \boldsymbol{N_p}\in{\mathbb{R}^{2}}$, and $\boldsymbol{N}=\begin{bmatrix}\boldsymbol{N_p}; &  \boldsymbol{N_r}\end{bmatrix}$. $\boldsymbol{{q}_p}=\begin{bmatrix}
\alpha & x\end{bmatrix}^T$, $\boldsymbol{{q}_r}=\begin{bmatrix}
\beta & q_r\end{bmatrix}^T$.

In this paper, we build the HTSMC based on \eqref{eq6} to realise spherical robot's direction control. Velocity controller have been proposed in our previous work in \cite{pre1}. 

\section{CONTROL ALGORITHM}
\begin{figure}[tb]
\centering
\includegraphics[width=7cm]{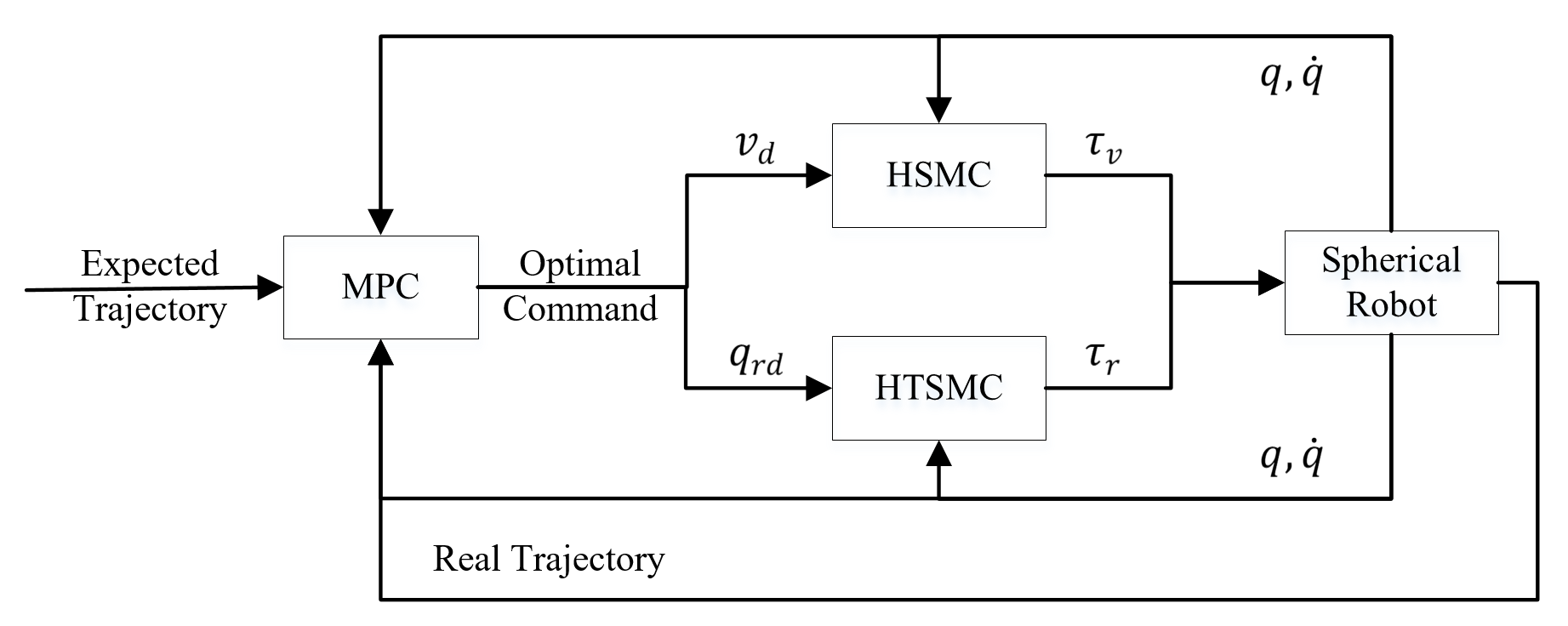}
\caption{The control scheme of MHH.}
\label{fig3}
\vspace{-0.5cm}
\end{figure}

In this section, we first introduce a direction controller named hierarchical terminal sliding mode controller (HTSMC) for spherical robot in subsection A. This controller adjusts the robot's direction by controlling its roll angle. Then, as for the instruction planning controller, we designed an MPC controller that that uses the reference trajectory to figure out the optimal control instructions. This is described in subsection B. With the velocity controller HSMC proposed in our previous work \cite{pre1, pre2}, we finally built the trajectory tracking framework MPC-HTSMC-HSMC (MHH) for unstable non-holonomic spherical robot. The scheme of the control structure is shown in Fig.~\ref{fig3}. The algorithm will then be talked about in detail.

\subsection{Direction Controller: HTSMC}

According to \eqref{eq2}, spherical robot's direction can be adjusted by modifying its roll angle since velocity has been specified by the sum of the velocity vectors in the x and y directions. Therefore, the roll angle controller is considered to as the direction controller. Before designing the controller, we need to transform the \eqref{eq6} into state space form. The state space equations are shown as follows:
\begin{equation}
\begin{cases}
\dot{x_1}=x_2\\
\dot{x_2}=f_1\left(\boldsymbol{\chi}\right)+b_1\left(\boldsymbol{\chi}\right)\tau_r\\
\dot{x_3}=x_4\\
\dot{x_4}=f_2\left(\boldsymbol{\chi}\right)+b_2\left(\boldsymbol{\chi}\right)\tau_r\label{eq7}\\
\end{cases}
\end{equation}
where $\boldsymbol{\chi}=\begin{bmatrix}x_1 & x_2 & x_3 & x_4\end{bmatrix}^T=\begin{bmatrix}\beta & \dot{\beta} & q_r & \dot{q}_r\end{bmatrix}^T$ is the state variable of the system, and
\begin{equation}
\begin{cases}
\begin{bmatrix}{f_1}\left(\boldsymbol{\chi}\right)&{f_2}\left(\boldsymbol{\chi}\right)\end{bmatrix}^T&=-{\boldsymbol{M_r}}^{-1}\cdot{\boldsymbol{N_r}}\\
\begin{bmatrix}{b_1}\left(\boldsymbol{\chi}\right)&{b_2}\left(\boldsymbol{\chi}\right)\end{bmatrix}^T&={\boldsymbol{M_r}}^{-1}\cdot\begin{bmatrix}1 & 1\end{bmatrix}^T
\notag.
\end{cases}
\end{equation}

Obviously it is a complex second-order control system, with the control purpose of making $q_r-q_{rd}=0$ where $q_{rd}$ is the target roll angle. Additionally, the system is also an SIMO one, and we adjust $q_r$ while influencing $\beta$. States associated with $\beta$ may cause instability in the control process if not taken into consideration, thus we have to add constraint for $\beta$. Under these conditions, the hierarchical terminal sliding mode controller (HTSMC) is proposed for this system. HTSMC has two layer sliding surfaces. The first layer sliding surface is combined with two subsystem sliding surfaces, respectively are $S_1\left(t\right)$ and $S_2\left(t\right)$. $S_1$ is also called the control surface which aims to accomplish the control purpose. The equation of $S_1$ is shown in \eqref{eq8}, and it is a kind of fast terminal SMC which can shorten time in sliding phase.
\begin{equation}
S_1=\dot{e}_{r}+a_1e_{r}+c_1e_{r}^{\frac{p_1}{q_1}}
\label{eq8}
\end{equation}
where $e_r=q_r-q_{rd}$, and $a_1, c_1, p_1, q_1$ are control parameters. And $0<p_1<q_1$. 

$S_2$ is called the stability surface to keep the stability by taking $\beta$'s control into consideration. When designing $S_2$, we found the heavy pendulum swing $\beta$ about the roll axis to balance lateral friction according to spherical robot's physical characteristics. Then we can get target pendulum swing angle $\beta_d$ as follows:
\begin{equation}
\beta_d = \arcsin{\frac{F_{fr}R}{mgl\cos{\alpha}}}
\label{eq9}
\end{equation}

$F_{fr}$ is the static friction since the robot moves without slipping and cannot been acquired. However, we can estimate the terminal $\hat{F}_{fr}$ according to $q_{rd}$ and the kinematic model. Turning radius $R_{turn}$ can be obtained when the roll angle has been adjusted to $q_{rd}$. And at this time $\hat{F}_{fr}$ is all applied to provide the centripetal force. Then we can get equations as follows:
\begin{equation}
R_{turn}=\frac{R}{\tan{q_r}}
\label{eq10}
\end{equation}
\begin{equation}
\hat{F}_{fr}=\frac{Mv^2}{R_{turn}}
\label{eq11}
\end{equation}

Replacing ${F}_{fr}$ with $\hat{F}_{fr}$, combining equations in \eqref{eq9}, \eqref{eq10} and \eqref{eq11}, then we can get $\beta_d$ as follows:
\begin{equation}
\beta_d = \arcsin{\frac{Mv^2\tan{q_r}}{mgl\cos{\alpha}}}
\label{eq12}
\end{equation}

When $q_r$ is adjusted to $q_{rd}$, $\beta$ should equal to $\beta_d$. After finding this constraint, we can design the stability surface $S_2$ as follows:
\begin{equation}
S_2=\dot{e}_{\beta}+a_2e_{\beta}+c_2e_{\beta}^{\frac{p_2}{q_2}}
\label{eq13}
\end{equation}
where $e_{\beta}=\beta-\beta_d$, and $a_2, c_2, p_2, q_2$ are control parameters. And $0<p_2<q_2$. 

The second layer sliding surface $S\left(t\right)$ can be seen as the combination of the first layer terminal sliding surfaces. The expression is shown as follows:
\begin{equation}
S=A\cdot S_1+B\cdot S_2
\label{eq14}
\end{equation}
where $A > B > 0$ are weight scale coefficients of the two subsystem surfaces. According to \eqref{eq7}, \eqref{eq8} and \eqref{eq13}, we can get the derivative of the second layer sliding surface as follows.
\begin{equation}
\begin{aligned}
\dot{S} &= A\cdot \dot{S}_1+B\cdot \dot{S}_2 \\
        &= A({f_2}-\ddot{q}_{rd}) +B({f_1}-\ddot{\beta}_d)+(A{b_2}+B{b_1})\tau_r+A{a_1}\dot{e}_r\\
        &\;\;\;\;+B{a_2}\dot{e}_{\beta}+Ac_1\frac{p_1}{q_1}e_r^{1-\frac{q_1}{p_1}}\dot{e}_r+Bc_2\frac{p_2}{q_2}e_\beta^{1-\frac{q_2}{p_2}}\dot{e}_\beta
\end{aligned}
\label{eq15}
\end{equation}

To drive the system to reach the sliding surface $S\left(t\right)=0$ as soon as possible, we choose the same reaching law as the velocity controller HSMC which is shown in \eqref{eq16}. The reaching law depends the control process and reaching speed, and also can be used to estimate the sliding surface in the next time.
\begin{equation}
\dot{S} = -k\cdot S-\varepsilon\cdot \text{sgn}\left(S\right),\label{eq16}
\end{equation}
where $k$ and $\varepsilon$ are constant and exponential reaching parameters, respectively. Based on the equations in \eqref{eq15} and \eqref{eq16}, we can get the control law of the output torque $\tau_r$ as follows:
\begin{equation}
\begin{aligned}
\tau_r   
    &= \frac{-1}{\left(A{b_2}+B{b_1}\right)}\left.[A({f_2}-\ddot{q}_{rd}) +B({f_1}-\ddot{\beta}_d)+A{a_1}\dot{e}_r\right.\\
    &\;\;\;\;\left.+B{a_2}\dot{e}_{\beta}+Ac_1\frac{p_1}{q_1}e_r^{1-\frac{q_1}{p_1}}\dot{e}_r+Bc_2\frac{p_2}{q_2}e_\beta^{1-\frac{q_2}{p_2}}\dot{e}_\beta\right.\\
    &\;\;\;\;\left.+k\cdot S+\varepsilon\cdot\text{sgn}\left(S\right) \right]
\end{aligned}
\label{eq17}
\end{equation}

Stability analysis of the HTSMC is shown as below. Take Lyapunov function as follows:
\begin{equation}
L=\frac{1}{2}S^2
\label{eq18}
\end{equation}

Apply \eqref{eq17} to \eqref{eq15}, then the derivative of $L$ will be:
\begin{equation}
\begin{aligned}
\dot{L} &= S\dot{S}\\
        &= S\left[-k\cdot S-\varepsilon\cdot\text{sgn}\left(S\right)\right]\\
        &= -k\cdot S^2-\varepsilon\cdot \left|S\right|\leq0
\end{aligned}
\label{eq19}
\end{equation}

According to \eqref{eq19}, $\dot{L}\leq0$. And when $\dot{L}\equiv0$, $S\equiv0$. According to the Lasalle’s invariance principle, this system is asymptotically stable \cite{bib2}. 

\subsection{Instruction Planning Controller: MPC}
Instruction planning controller is necessary to tell the robot how to track the reference trajectory with minimum cost. Then two torque controllers will work to execute the instructions. In this subsection, an MPC controller is designed to find the optimal control policy. The cost function $J(\cdot)$ that has to be minimized to control the trajectory of the robot is in general given by the equation below:

\begin{align}
\min_{\boldsymbol{\overline{U}}(\cdot)} & && J(\cdot)=\sum_{i=1}^N\left\|\boldsymbol{\overline{X}}-\boldsymbol{{X}}_{ref}\right\|_{\boldsymbol{Q}}+\left\|\boldsymbol{\overline{U}}-\boldsymbol{{U}}_{ref}\right\|_{\boldsymbol{R}}  \label{eq20}\\
\text{s.t.} & &&
\boldsymbol{\dot{\overline{X}}}(i|k)=F\left(\boldsymbol{\overline{X}}(i|k), \boldsymbol{\overline{U}}(i|k)\right), \tag{20a} \label{eq20a}\\
& && \boldsymbol{{\overline{X}}}(0|k)=\boldsymbol{{\overline{X}}}_k,\;\;\boldsymbol{{\overline{U}}}(0|k)=\boldsymbol{{\overline{U}}}_k, \tag{20b} \label{eq20b}\\
& && \boldsymbol{{\overline{X}}}(i|k)\in\left[
\boldsymbol{\overline{X}}_{min},\;\; \boldsymbol{\overline{X}}_{max}\right], \tag{20c} \label{eq20c}\\
& && \boldsymbol{{\overline{U}}}(i|k)\in\left[
\boldsymbol{\overline{U}}_{min},\;\; \boldsymbol{\overline{U}}_{max}\right], \tag{20d} \label{eq20d} \\
& && G\left(\boldsymbol{\overline{X}}(i|k),\;\; \boldsymbol{\overline{U}}(i|k)\right)= 0 \tag{20e} \label{eq20e}
\end{align}
where $N$ is the prediction horizon, ${\boldsymbol{Q}}$ and ${\boldsymbol{R}}$ are definite weighting coefficient matrix, the states and inputs at moment $i+k$ are defined as $\boldsymbol{\overline{X}}=\begin{bmatrix} X & Y & \phi \end{bmatrix}^T$ and $\boldsymbol{\overline{U}}=\begin{bmatrix} v & q_r \end{bmatrix}^T$, respectively. $\boldsymbol{{X}}_{ref}=\begin{bmatrix} X_{ref} & Y_{ref} & \phi_{ref} \end{bmatrix}^T$ and $\boldsymbol{{U}}_{ref}=\begin{bmatrix} v_{ref} & q_{ref} \end{bmatrix}^T$ are reference trajectory and reference inputs, respectively. \eqref{eq20a} is the kinematic model which is also presented in \eqref{eq2}. Initial states and inputs are defined as $\boldsymbol{{\overline{X}}}_k$ and $\boldsymbol{{\overline{U}}}_k$, respectively, which are obtained by sensors. Limitations for states and inputs are respectively presented in \eqref{eq20c} and \eqref{eq20d}. And equality constraint \eqref{eq20e} is employed to indicate the boundary value problem of direct multi-shooting method.

After defining the optimization problem, we solve it by employing the Sequential Quadratic Programming (SQP) approach. The sequence of optimal control inputs $\boldsymbol{\overline{U}}(\cdot)$ is then obtained. Then we apply $\boldsymbol{\overline{U}}(1|k)$ to velocity controller HSMC and direction controller HTSMC. Following that, the two controllers execute the optimal commands in a predictable manner. Then we can get the optimal trajectory which is close to the reference trajectory. Finally the whole trajectory tracking framework MHH is established. 

Furthermore, to make sure the system is stable when tracking trajectories, MPC may need stability constraints like limits on angle velocity and acceleration. And dynamic model may also be required. This may make the optimal problem hard to solve. However, two Lyapunov-based sliding mode controllers ensures the stability of the system. This can eliminate the requirement for stability constraints for MPC, which can simplify the nonlinear program in \eqref{eq20}. Therefore, the trajectory tracking framework can be of stability.

\section{EXPERIMENTAL RESULTS}
In this section, we aims to verify the effectiveness and robustness of the direction controller (or named roll angle controller) HTSMC and trajectory tracking framework MHH. In subsection A, experiments are carried out to test the roll angle controller HTSMC's tracking effect. For comparison, the same experiments were conducted using the fuzzy-PID controller proposed in our previous work which is verified to be better than traditional PID controller \cite{pre3}. In subsection B, two trajectory tracking frameworks are compared by tracking 
three kinds of trajectories. One is the MHH (MPC-HSMC-HTSMC) and the other is MHF (MPC-HSMC-Fuzzy-PID) whose direction controller is fuzzy-PID rather than HTSMC. 

Furthermore, all of the experiments are carried out on the flat tiled floor, where whole-body dynamic model's parameters were obtained through system identification. Information about angles, velocities, and accelerations is obtained using an IMU and an encoder. Algorithms run in concurrent threads on a mini PC (Intel i7-8559U, 2.70 GHz, Quad-core 64-bit). The calculated output torque will be transmitted to the lower computer (TI TMS320F28069) to control the motors. Moreover, the sampling time is 20ms, control frequency is 50Hz for velocity controller HSMC and direction controller HTSMC, while the prediction frequency is 10Hz for MPC.

\subsection{Direction Control Experiments}

\subsubsection{Single Roll Angle Tracking Under Different Velocities}

\begin{figure}[tb]
\centering
\subfigure[When velocity is 0.5m/s]
{
    \label{fig:subfig:a} 
    \includegraphics[width=6cm]{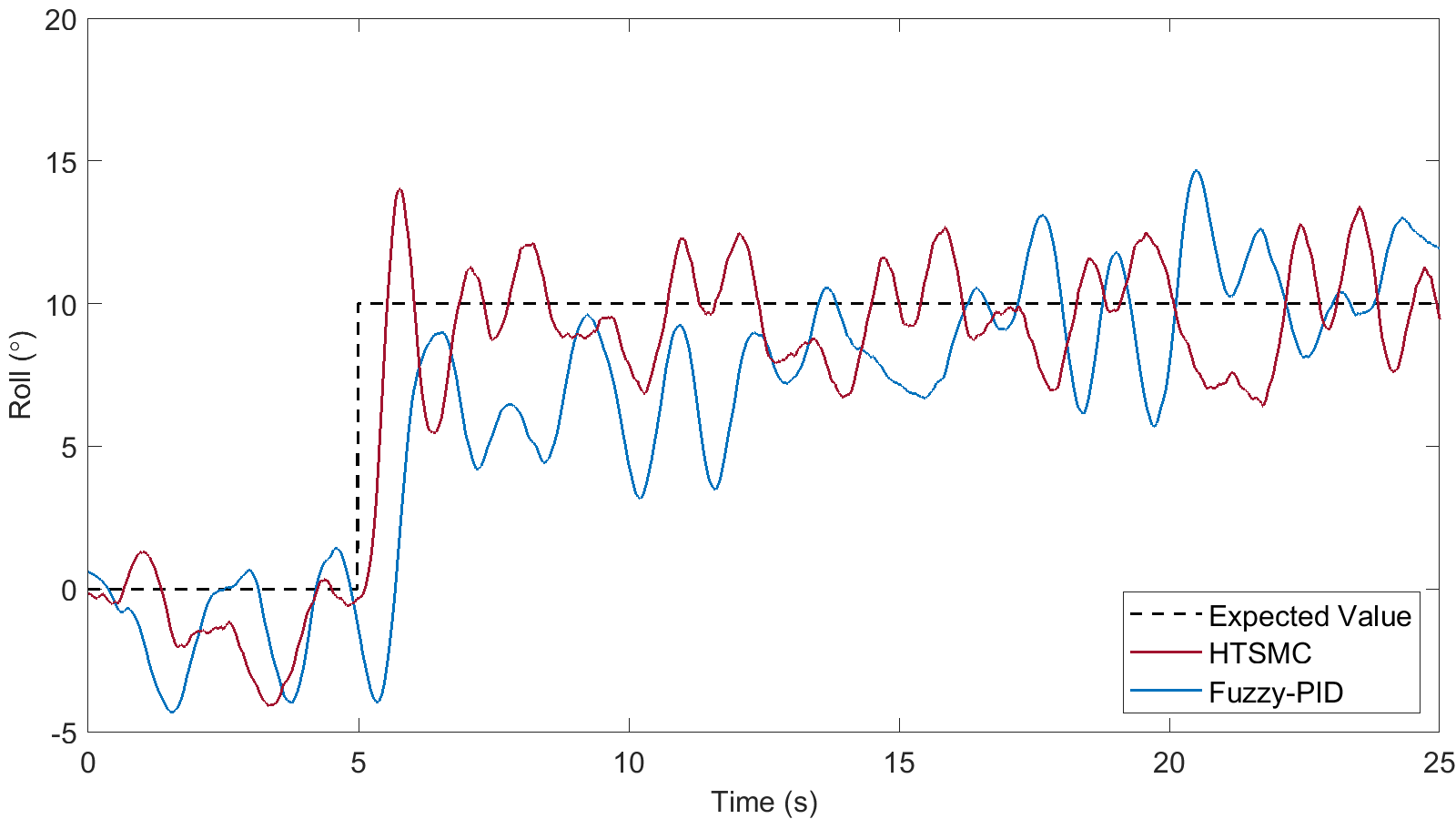}}
    \hspace{0in}    
\subfigure[When velocity is 1.0m/s]
{
    \label{fig:subfig:b} 
    \includegraphics[width=6cm]{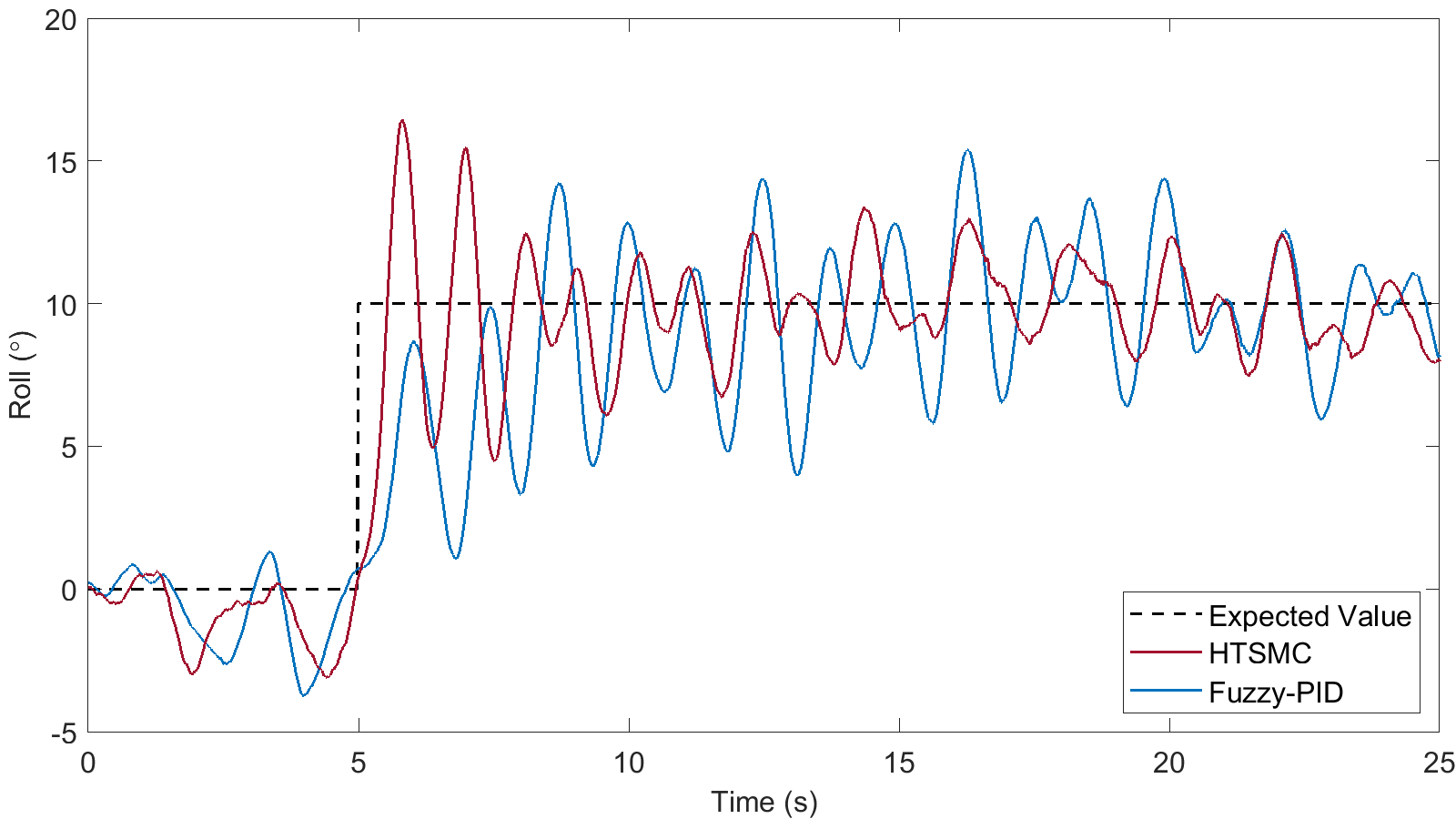}}
    \hspace{0in}

\caption{Single roll angle tracking.}
\label{fig_ex1}
\end{figure}

In this portion, our purpose is to turn the spherical robot left at velocities of 0.5m/s and 1.0m/s. As previously explained, turning motion can be achieved by controlling its roll angle. And in this experiments, we hope the robot can move straight for the first five seconds and then turn left immediately with the expected roll angle of ${10^\circ}$. The results are shown in Fig.~\ref{fig_ex1}.

In order to better compare the control effect of the two controllers, we selected five indicators, i.e. rising time $t_r$, setting time $t_s$, mean error $e_{m}$ which is the average error in the turning process (from 5s to the end), root mean square error value of the roll angle $e_{rmse}$ in steady-state process and the error boundary $\Delta h$. We assume the rising time is the time when the system first reaches the expected value. And steady-state is reached when envelope of the angle starts to remain basically horizontal. Results are shown in Table.~\ref{table2}.
\begin{table}[tb]
\begin{center}
\linespread{1.2}
\caption{Single Roll Angle Control Effects.}
\resizebox{.99\columnwidth}{!}{
\begin{tabular}{ccccccc}
\toprule
\multirow{2}*{\textbf{Velocity}} & \multirow{2}*{\textbf{Controller}} &  \multicolumn{5}{c}{\textbf{Indicators}}\\
\cmidrule(lr){3-7}
 & &\textbf{\textit{$\boldsymbol{t_r}$(s)}}&\textbf{\textit{$\boldsymbol{t_s}$ (s)}} & \textbf{\textit{$\boldsymbol{e_{m}}$ (m/s)}} &\textbf{\textit{$\boldsymbol{e_{rmse}}$ (m/s)}}
 &\textbf{\textit{$\boldsymbol{\Delta h}$($^\circ$)}}\\
\midrule
\multirow{2}{*}{\tabincell{c}{When 0.5m/s}} & HTSMC & 0.54  & 1.61 & -0.358  & 1.727 & -3.59, +3.38\\
 & Fuzzy-PID      & 8.51  & 6.86 & -0.351 & 2.097 & -6.28, +4.61\\
\cmidrule(lr){2-7}
\multirow{2}{*}{When 1.0m/s} & HTSMC & 0.54  & 2.68 & -0.097 & 1.509  & -3.88, +3.39\\
 & Fuzzy-PID      & 3.41  & 3.66 & -0.172 & 2.5373 & -5.968, +5.32 \\
\bottomrule
\end{tabular}}
\label{table2}
\end{center}
\end{table}

According to results in Fig.~\ref{fig_ex1} and Table.~\ref{table2}, it is obvious that HTSMC has much smaller rising time and setting time compared with Fuzzy-PID, which means it can reach the setting point very fast, able to return stable and steady state quickly. This is owing to the design of its hierarchical sliding surfaces. We also tried to increase the $k_p$ or $k_i$ of Fuzzy-PID, but the outcome was system oscillation and it was difficult to reestablish stability. On the other hand, results also showed that HTSMC has higher stability than Fuzzy-PID, since it has much smaller $e_{rmse}$ and $\Delta h$. It is also proved that our new controller has similar good performance under different velocities, which means high robustness. 

\subsubsection{Multiple Discrete Roll Angle Tracking}
\begin{figure}[tb]
\centering
\includegraphics[width=6cm]{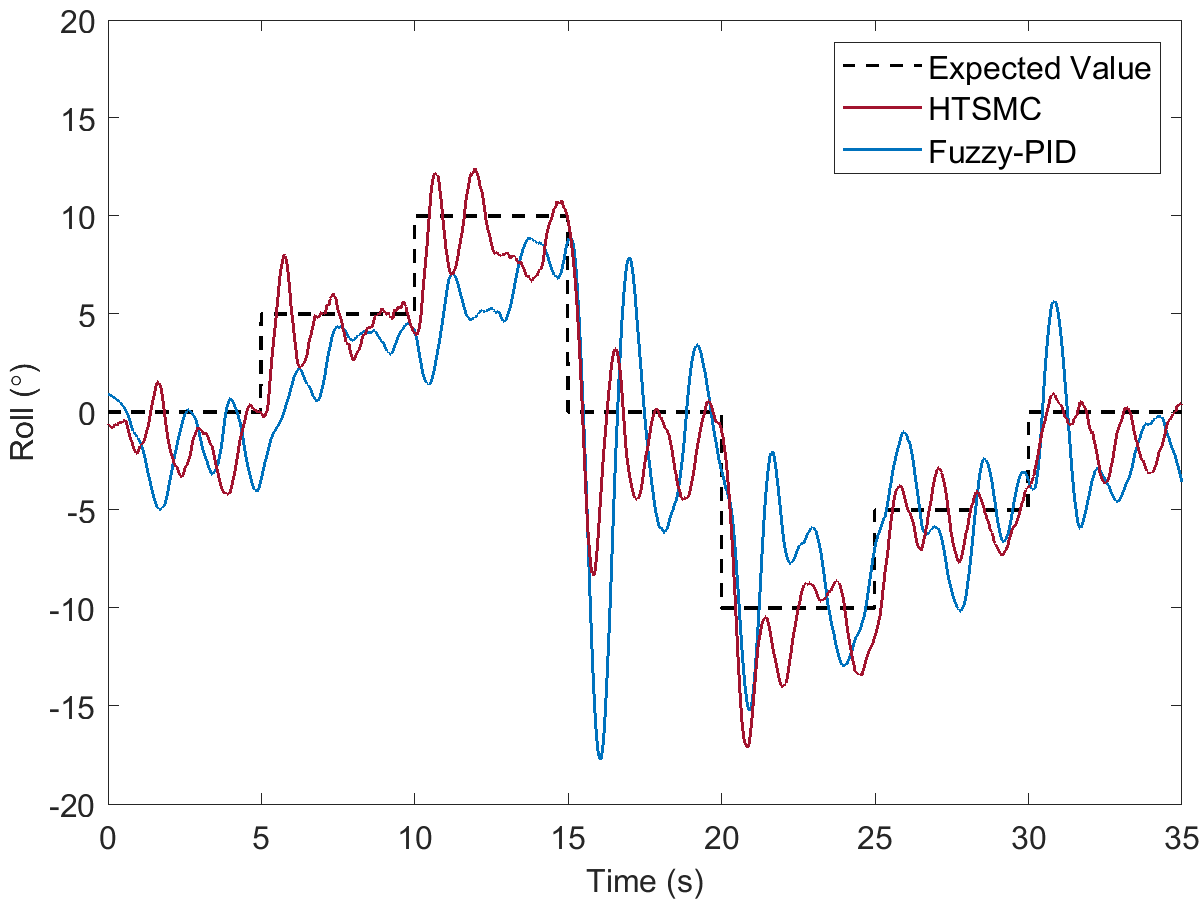}
\caption{Multiple discrete roll angle tracking.}
\label{fig_ex2}
\end{figure}

In this test, we would like to operate the spherical robot to track a time-variant roll angle which is a discrete function. The expected roll angle changes every 5 seconds. HSMC controls the robot to keep a velocity of 0.5m/s. The results can be seen in Fig.~\ref{fig_ex2}.

In order to further compare the control performance of the two controllers, we choose their root mean square error value $e_{rmse}$ and mean absolute error $e_{mae}$ of the tracking roll angle in the whole procedure to indicate the tracking effect. Results of the two indicators are shown in Table.~\ref{table3}.
\begin{table}[tb]
\begin{center}
\linespread{1.2}
\caption{Time-variant Angle Tracking.}
\resizebox{.7\columnwidth}{!}{
\begin{tabular}{cccc}
\toprule
\multirow{2}*{\textbf{Velocity}} & \multirow{2}*{\textbf{Controller}} &  \multicolumn{2}{c}{\textbf{Indicators}}\\
\cmidrule(lr){3-4}
 &  & \textbf{\textit{$\boldsymbol{e_{mae}}$ (m/s)}} &\textbf{\textit{$\boldsymbol{e_{rmse}}$ (m/s)}}\\
\midrule
\multirow{2}{*}{\tabincell{c}{Discrete Tracking}} & HTSMC & 1.852  & 2.511\\
 & Fuzzy-PID      & 3.191  & 4.128\\
\cmidrule(lr){2-4}
\multirow{2}{*}{Sine-wave Tracking} & HTSMC & 1.631  & 2.082\\
 & Fuzzy-PID      & 2.115  & 2.600\\
\bottomrule
\end{tabular}}
\label{table3}
\end{center}
\vspace{-0.6cm}
\end{table}

According to results in Fig.~\ref{fig_ex2} and Table.~\ref{table3}, it is obvious that HTSMC offers superior tracking performance and fewer errors than Fuzzy-PID. And the HTSMC controller offers a quicker response time and a shorter rising time. HTSMC exhibits a similar control effect and robustness for all expected angles in this experiment, while the other one does not. Specifically, the data imply that the HTSMC has higher stability in discrete roll angle tracking.
\begin{figure}[tb]
\centering
\includegraphics[width=6cm]{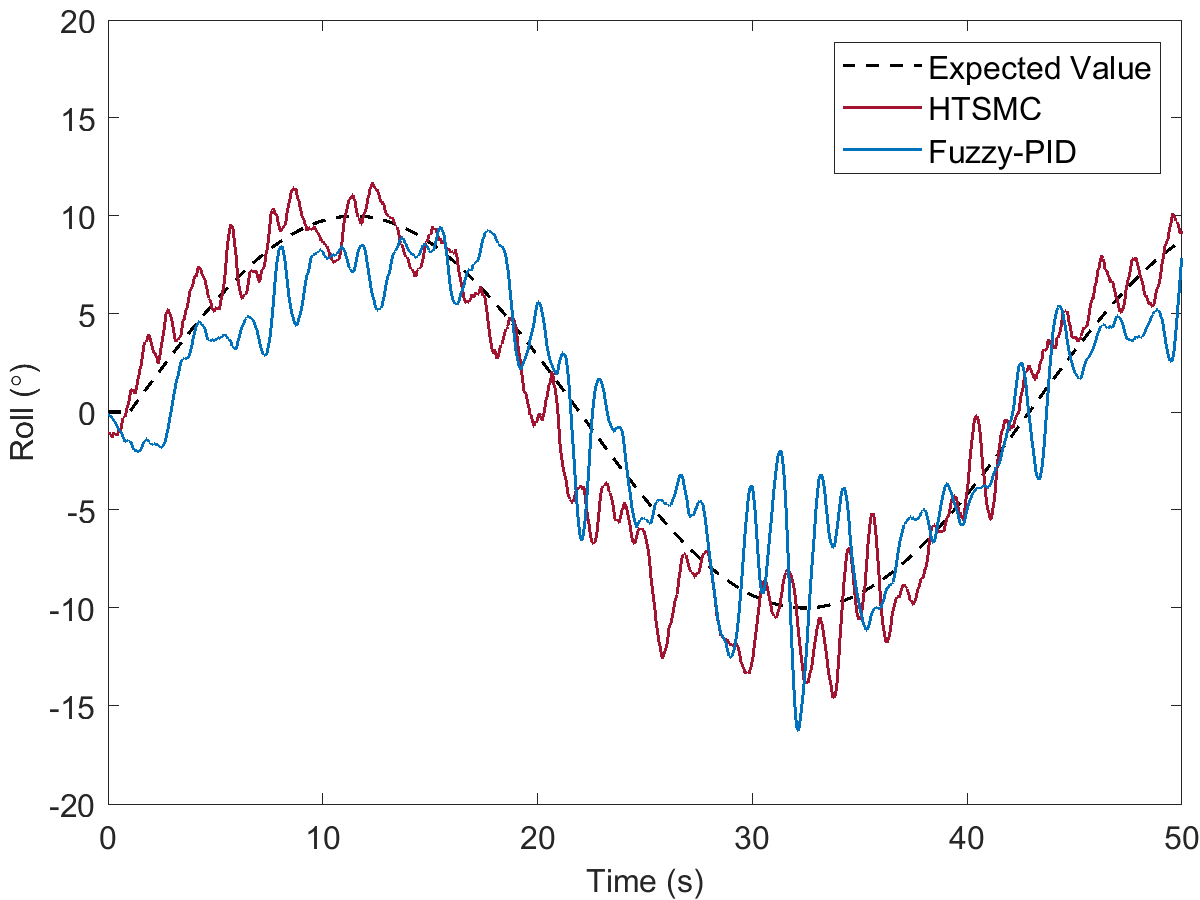}
\caption{Sine-wave tracking experiment.}
\label{fig_ex3}
\end{figure}

\begin{figure*}[tb]
\centering
\subfigure[Real Path of MHH]
{
    \label{fig:subfig:a}
    \includegraphics[height=2.2cm]{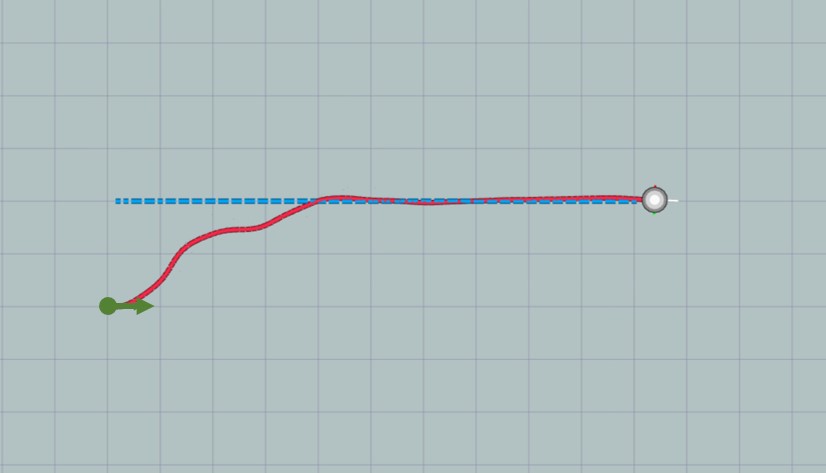}}
    \hspace{0in}    
\subfigure[Real Path of MHF]
{
    \label{fig:subfig:b}
    \includegraphics[height=2.2cm]{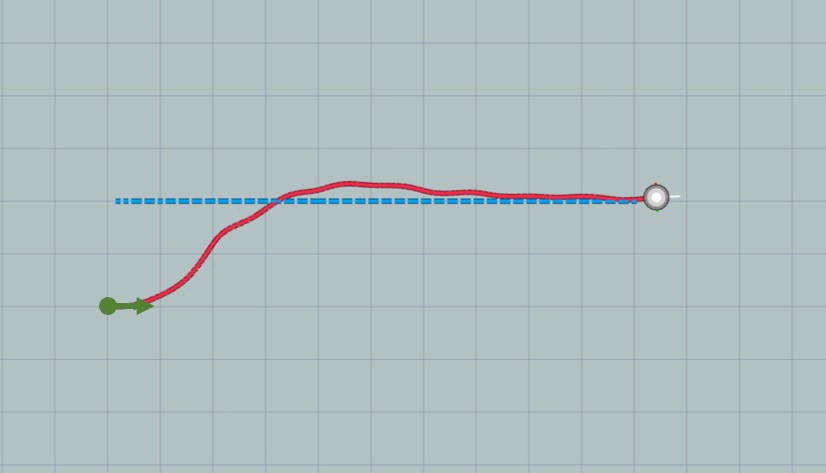}}
    \hspace{0in}
\subfigure[Distance]
{
    \label{fig:subfig:c}
    \includegraphics[height=2.2cm]{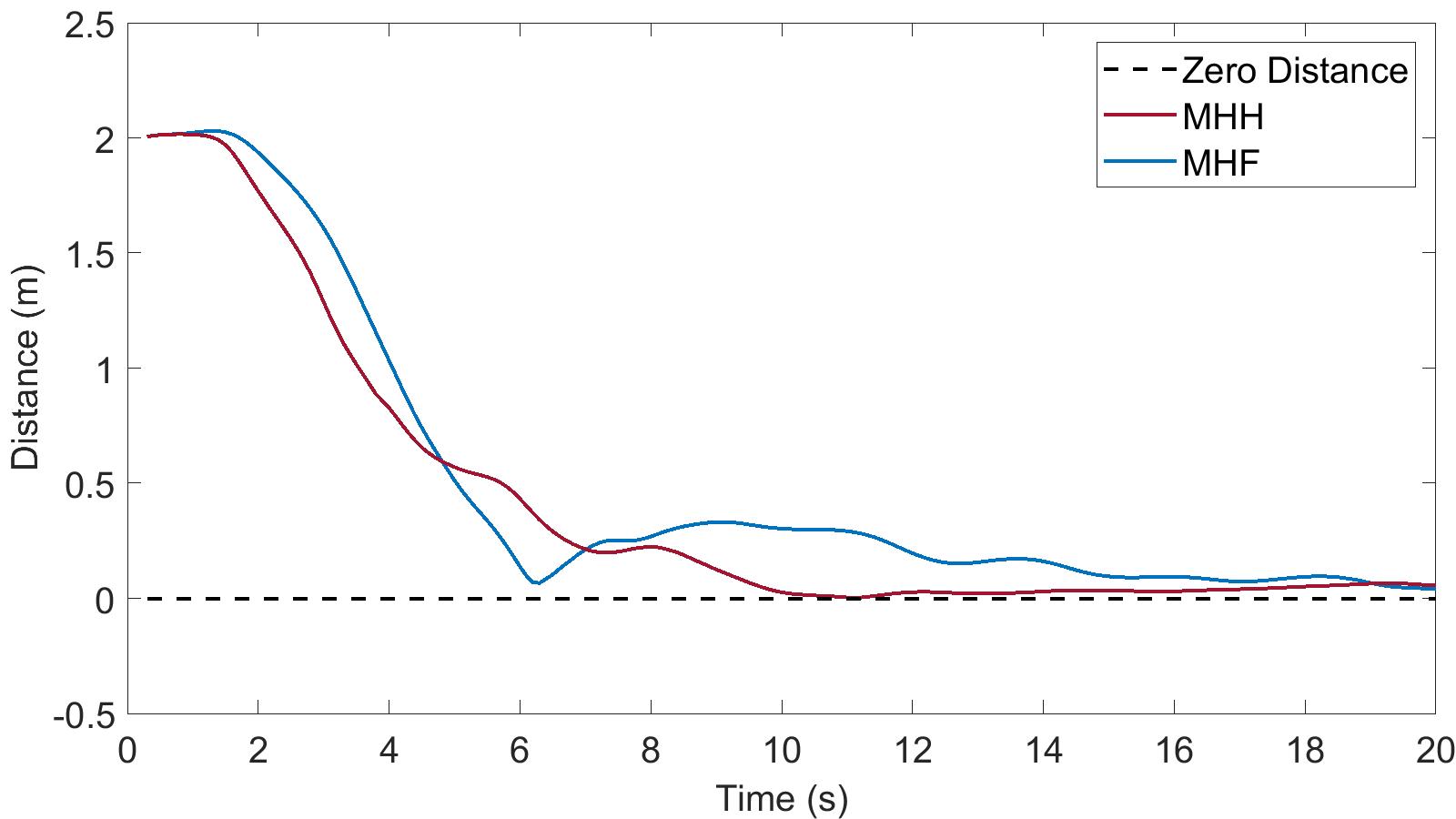}}
    \hspace{0in}\\
\subfigure[Tracking of $X$]
{
    \label{fig:subfig:d}
    \includegraphics[height=2.2cm]{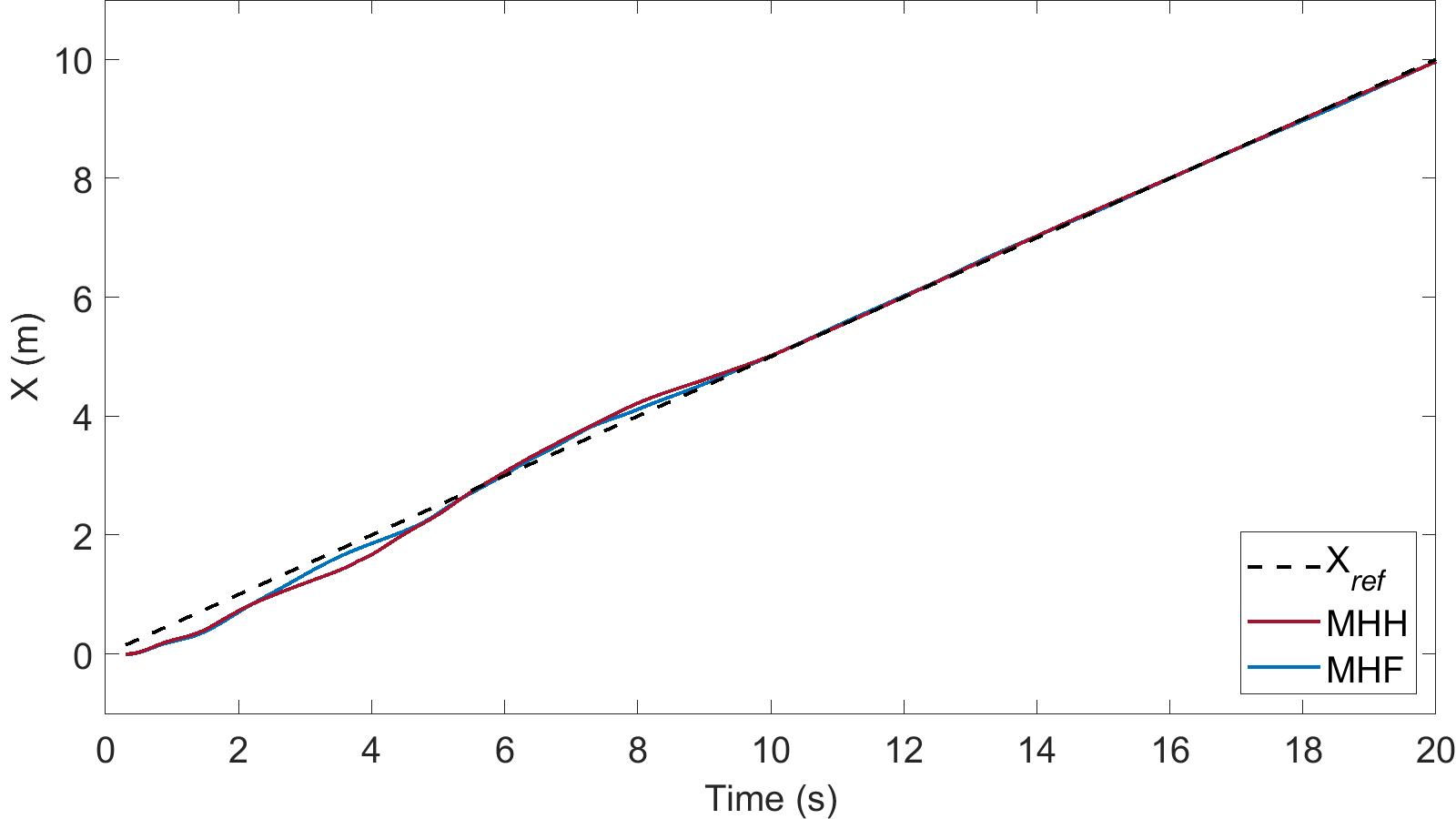}}
    \hspace{0in}
\subfigure[Tracking of $Y$]
{
    \label{fig:subfig:e}
    \includegraphics[height=2.2cm]{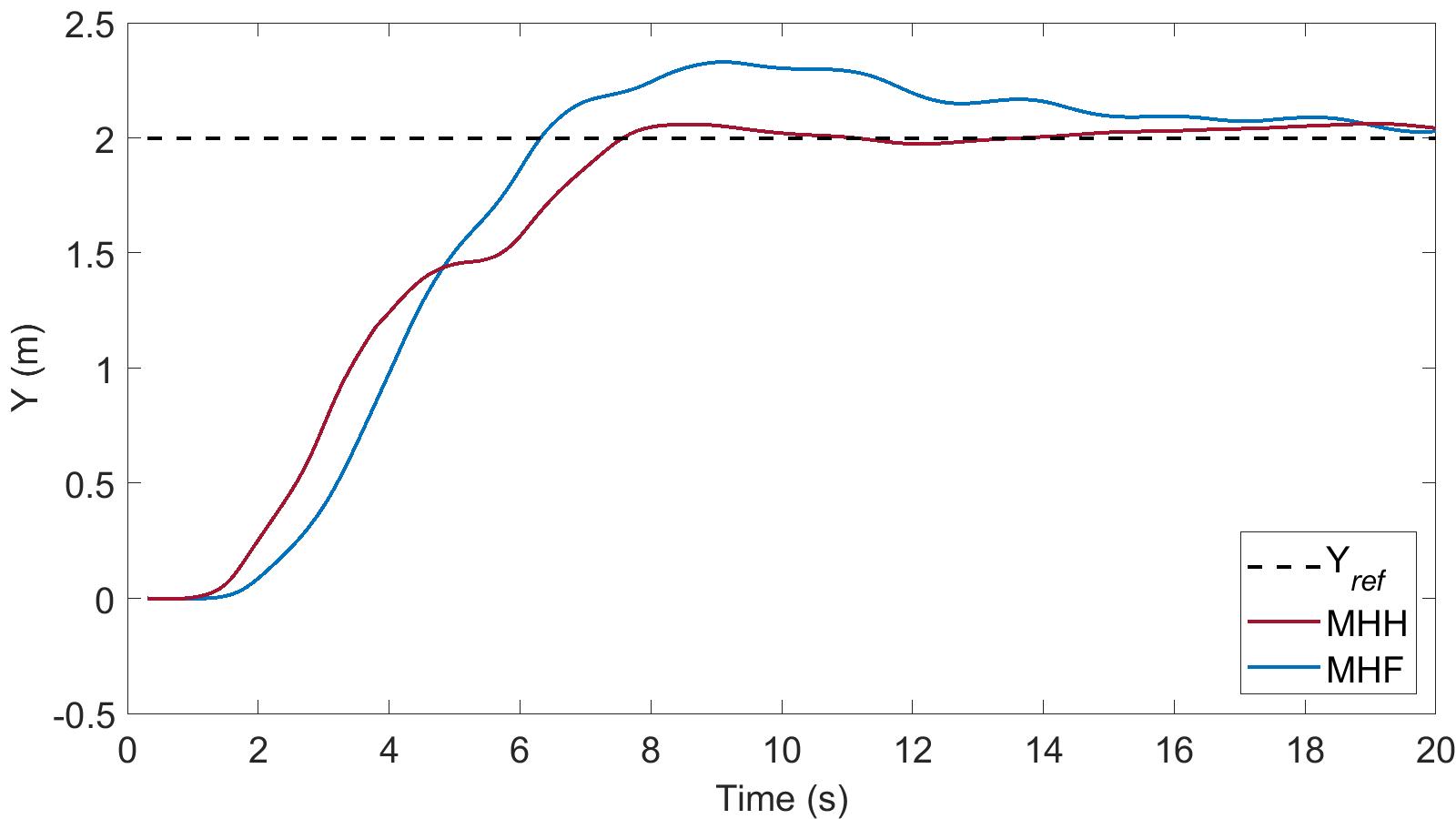}}
    \hspace{0in}
\subfigure[Tracking of $\phi$]
{
    \label{fig:subfig:f}
    \includegraphics[height=2.2cm]{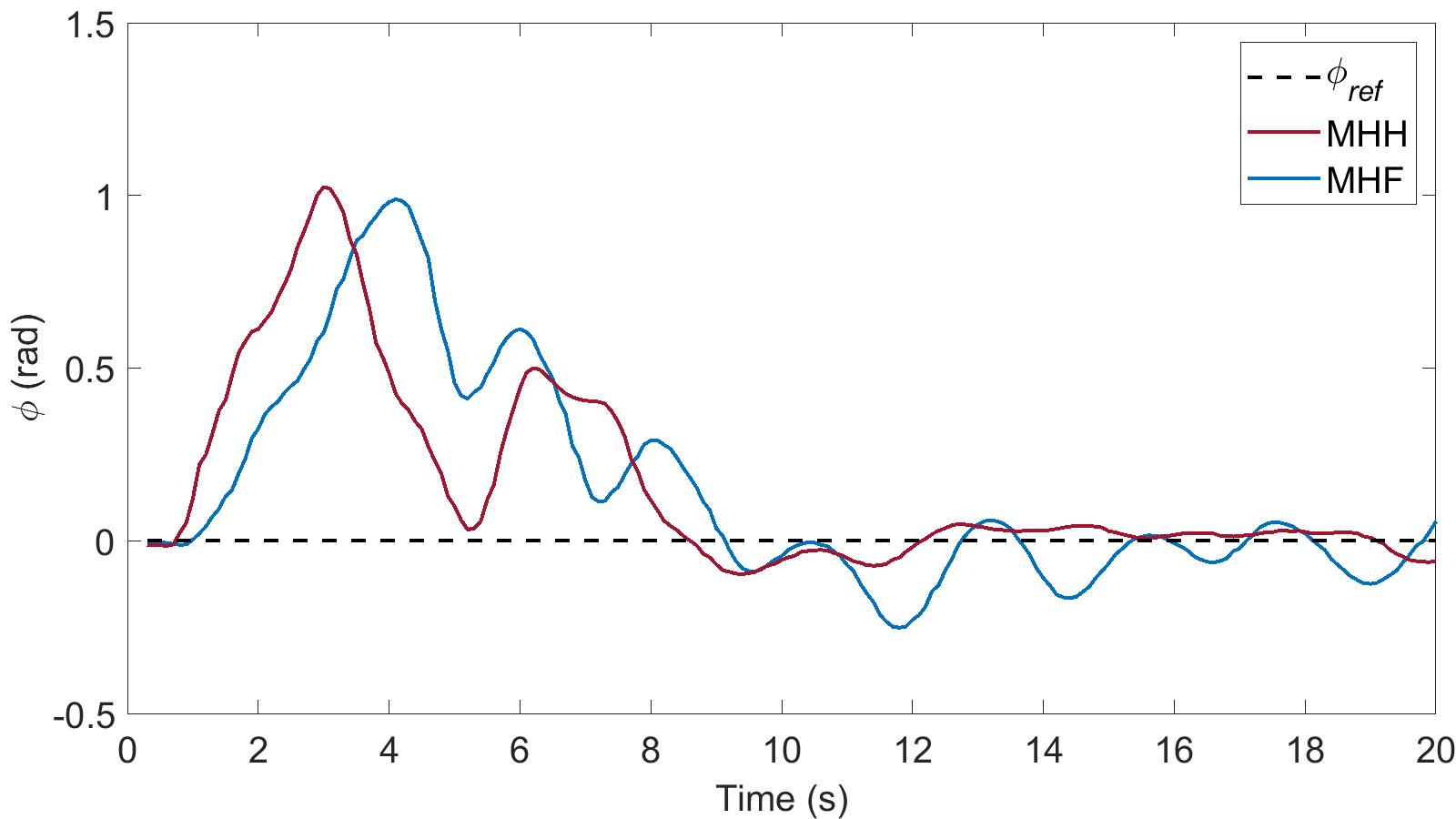}}
\caption{Tracking the trajectory of a line.}
\label{Traj_1}
\end{figure*}

\begin{figure*}[htb]
\centering
\subfigure[Real Path of MHH]
{
    \label{fig:subfig:a}
    \includegraphics[height=2.2cm]{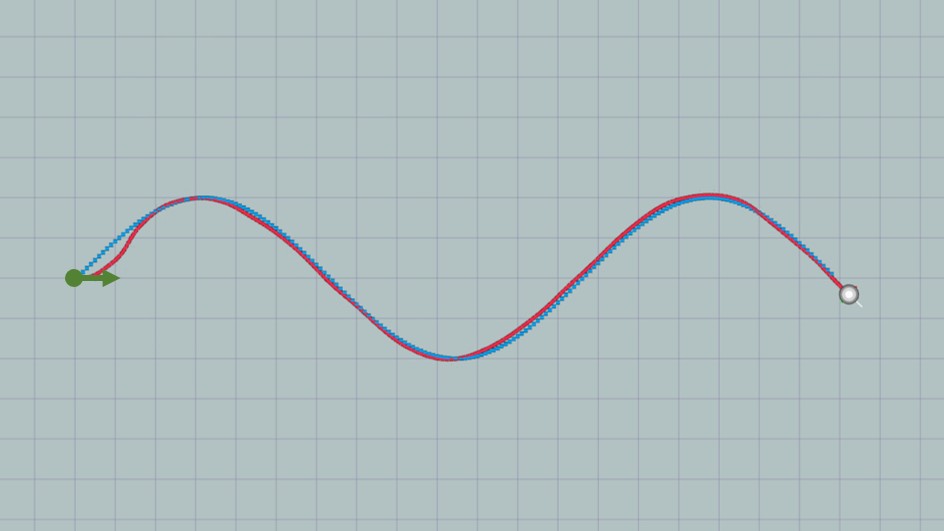}}
    \hspace{0in}    
\subfigure[Real Path of MHF]
{
    \label{fig:subfig:b}
    \includegraphics[height=2.2cm]{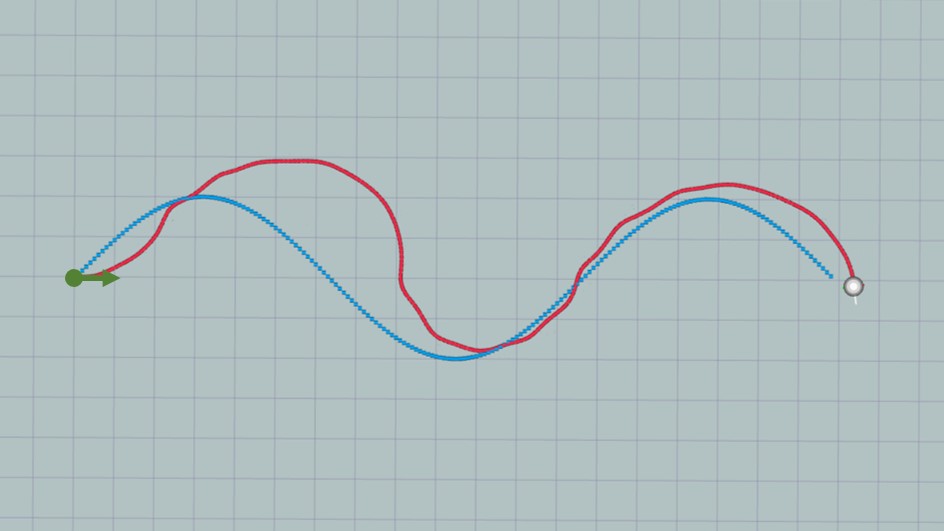}}
    \hspace{0in}
\subfigure[Distance]
{
    \label{fig:subfig:c}
    \includegraphics[height=2.2cm]{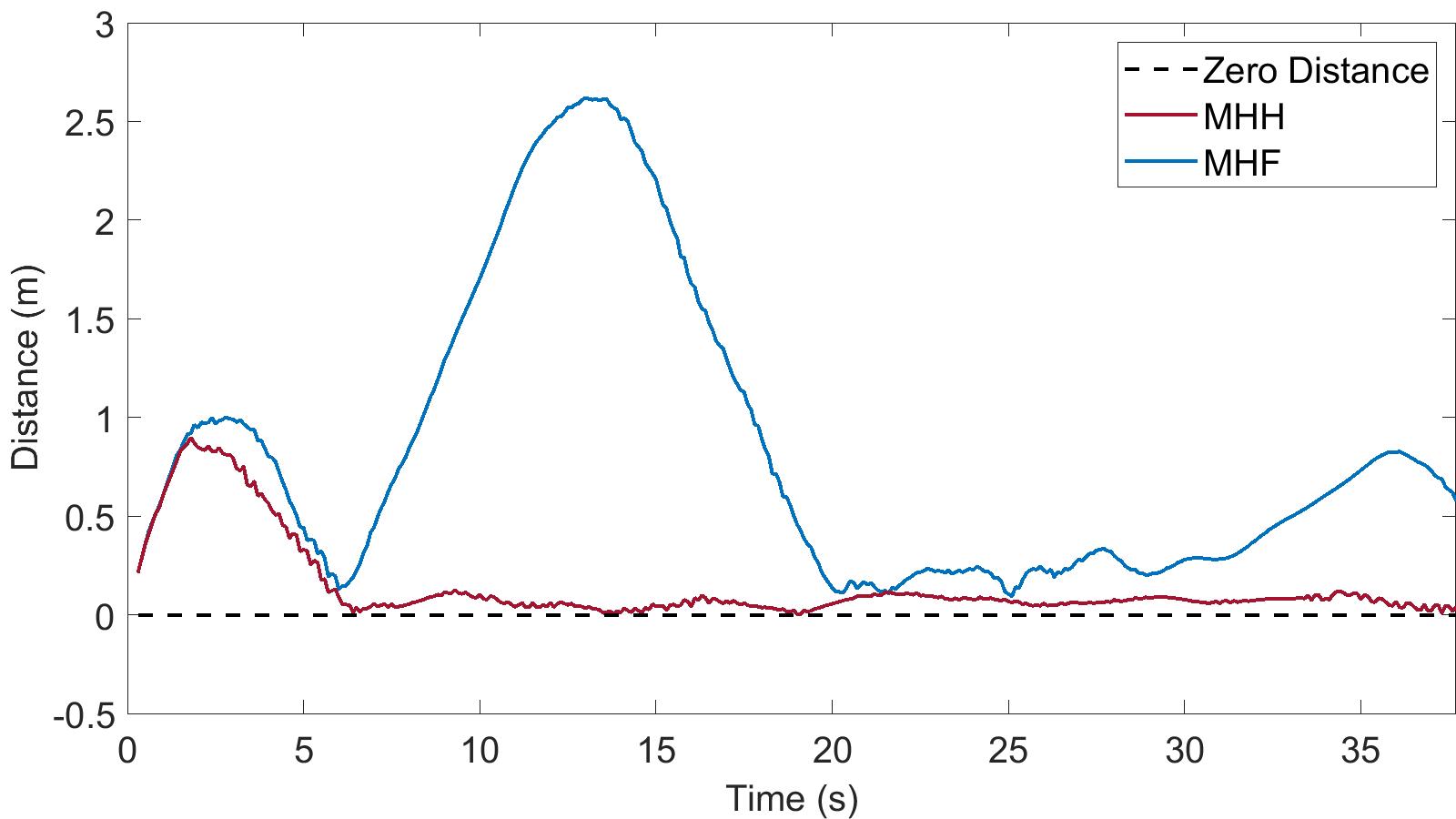}}
    \hspace{0in}\\
\subfigure[Tracking of $X$]
{
    \label{fig:subfig:d}
    \includegraphics[height=2.2cm]{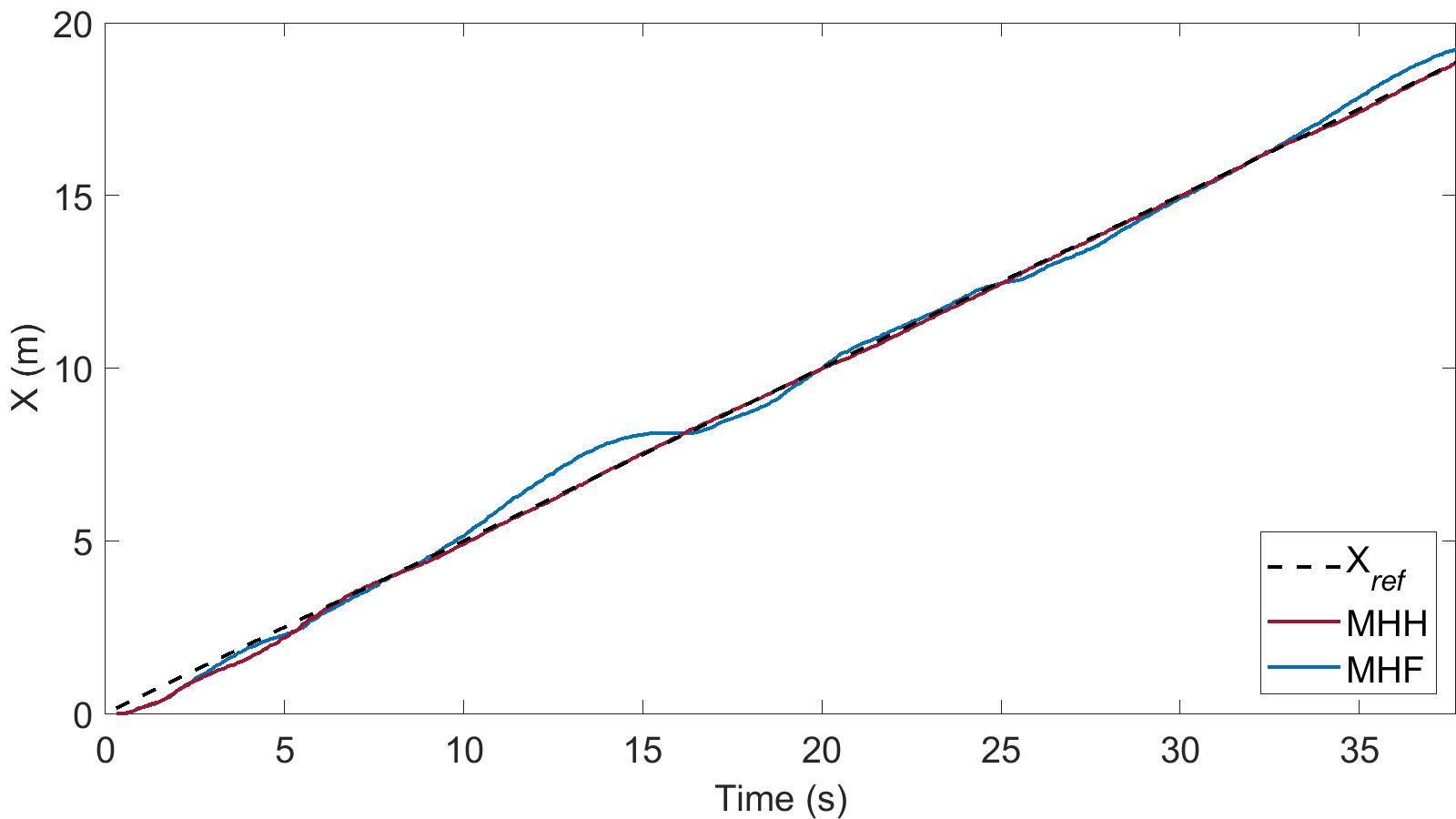}}
    \hspace{0in}
\subfigure[Tracking of $Y$]
{
    \label{fig:subfig:e}
    \includegraphics[height=2.2cm]{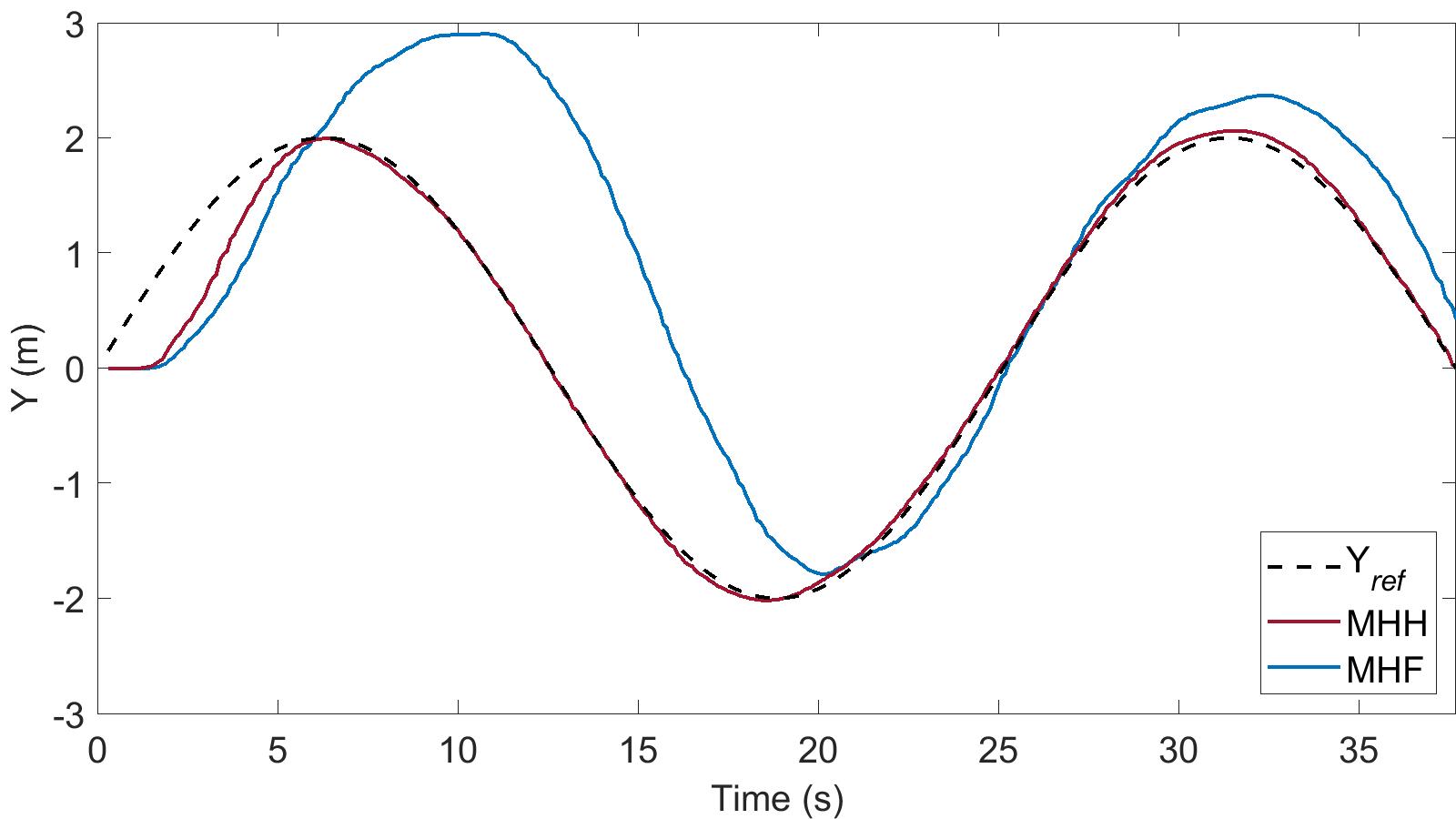}}
    \hspace{0in}
\subfigure[Tracking of $\phi$]
{
    \label{fig:subfig:f}
    \includegraphics[height=2.2cm]{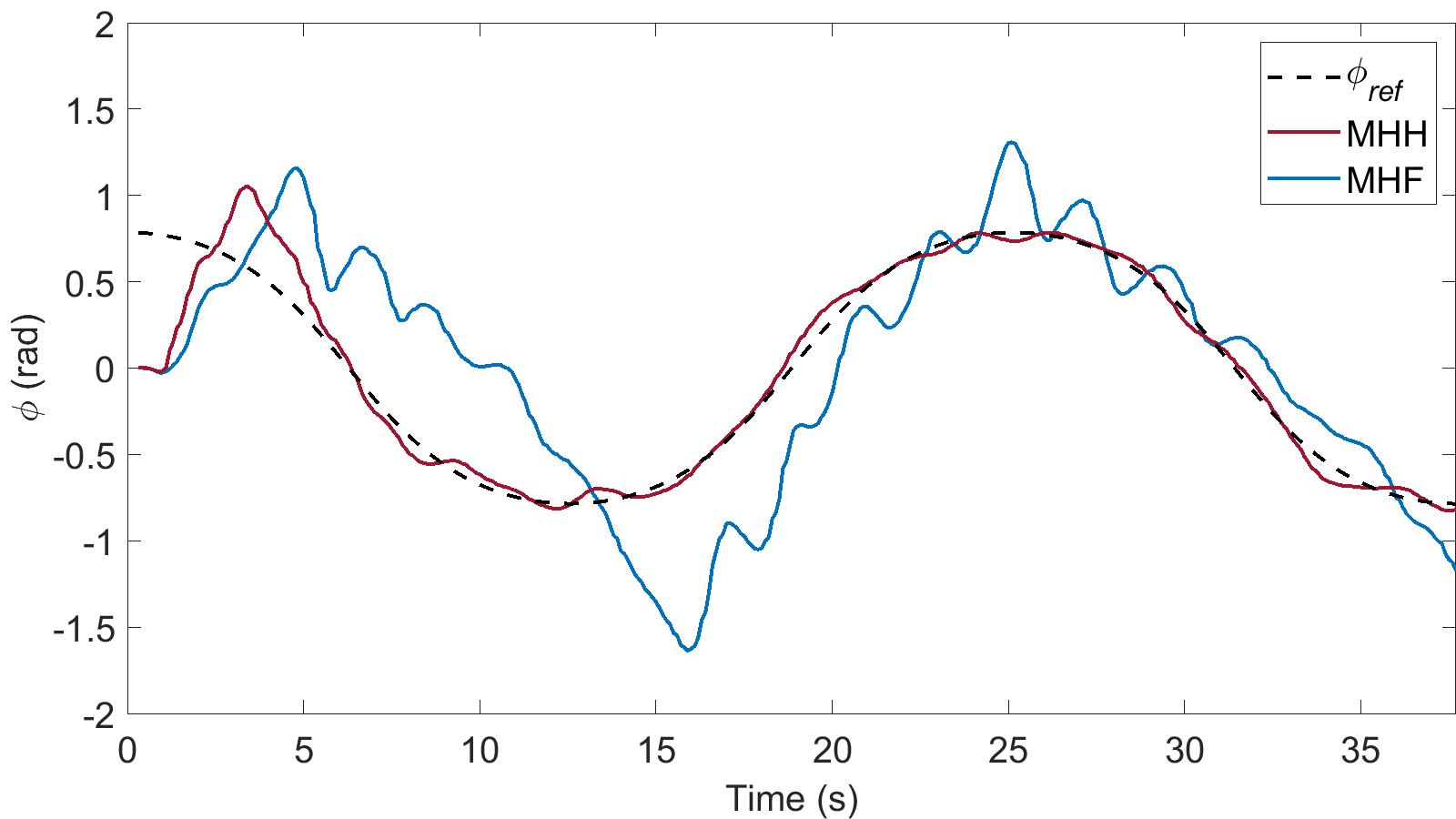}}
\caption{Tracking the trajectory of a sine-wave.}
\label{Traj_2}
\end{figure*}

\begin{figure*}[htb]
\centering
\subfigure[Real Path of MHH]
{
    \label{fig:subfig:a}
    \includegraphics[height=2.2cm]{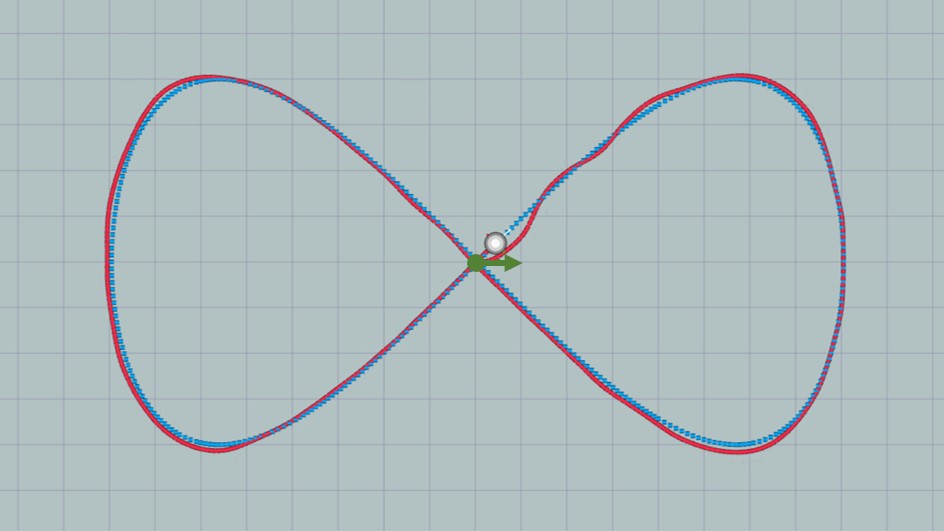}}
    \hspace{0in}    
\subfigure[Real Path of MHF]
{
    \label{fig:subfig:b}
    \includegraphics[height=2.2cm]{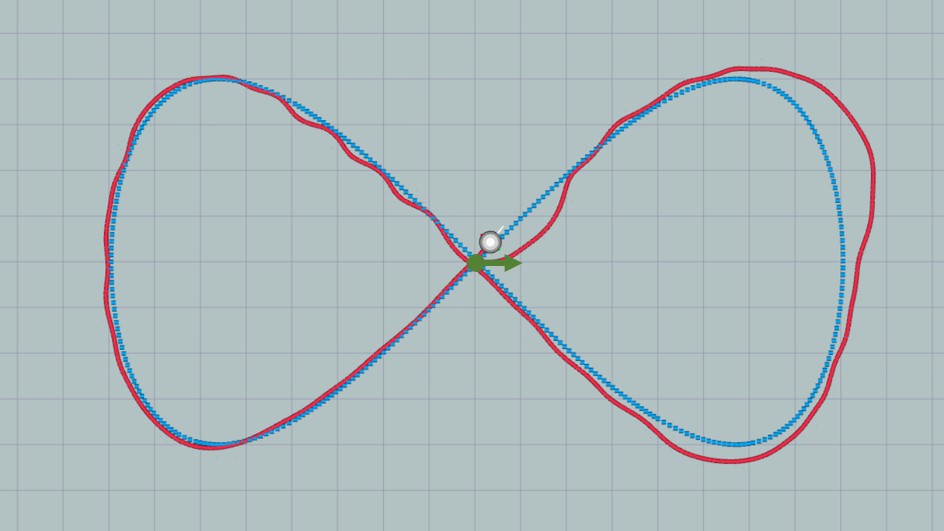}}
    \hspace{0in}
\subfigure[Distance]
{
    \label{fig:subfig:c}
    \includegraphics[height=2.2cm]{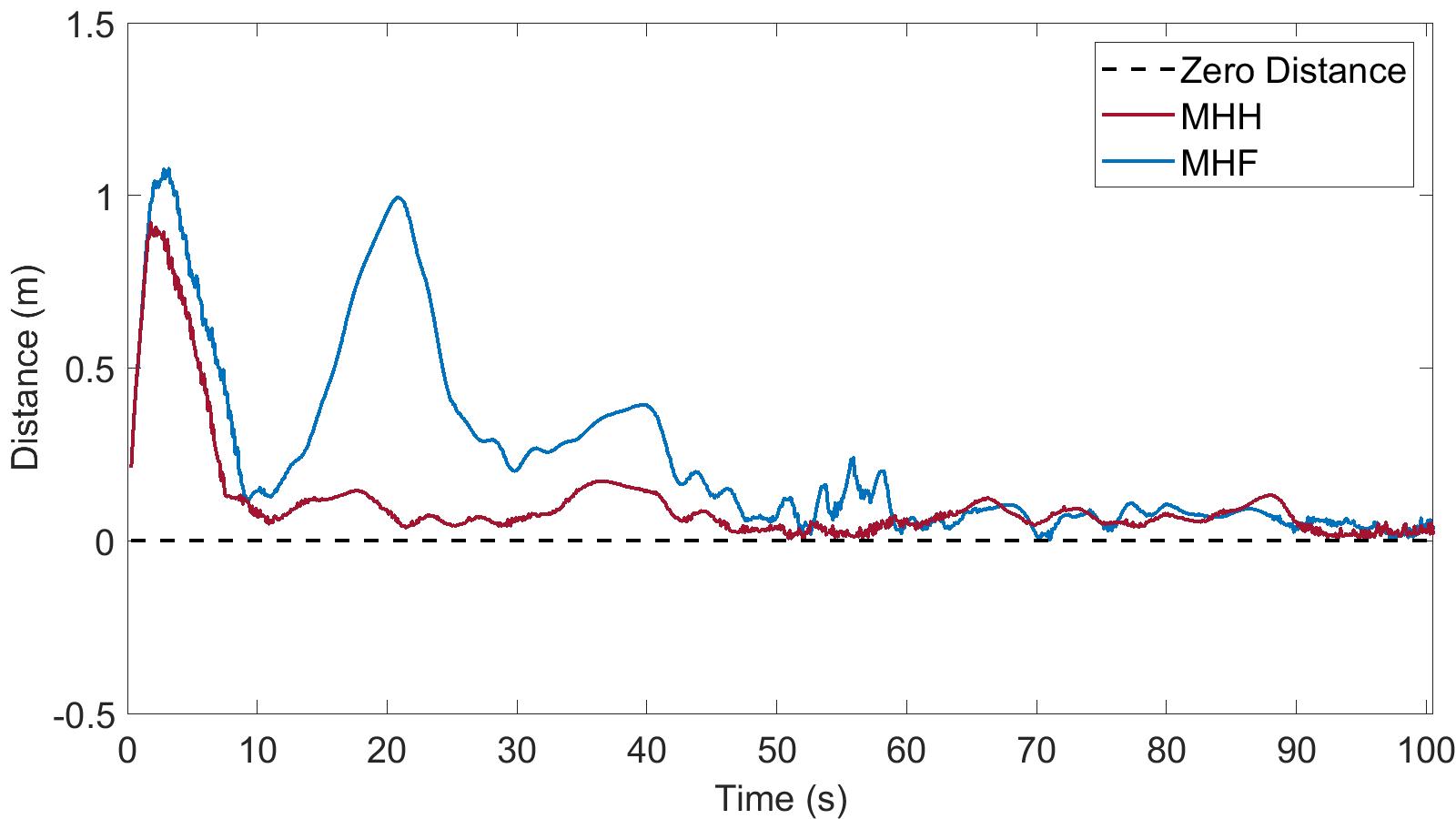}}
    \hspace{0in}\\
\subfigure[Tracking of $X$]
{
    \label{fig:subfig:d}
    \includegraphics[height=2.2cm]{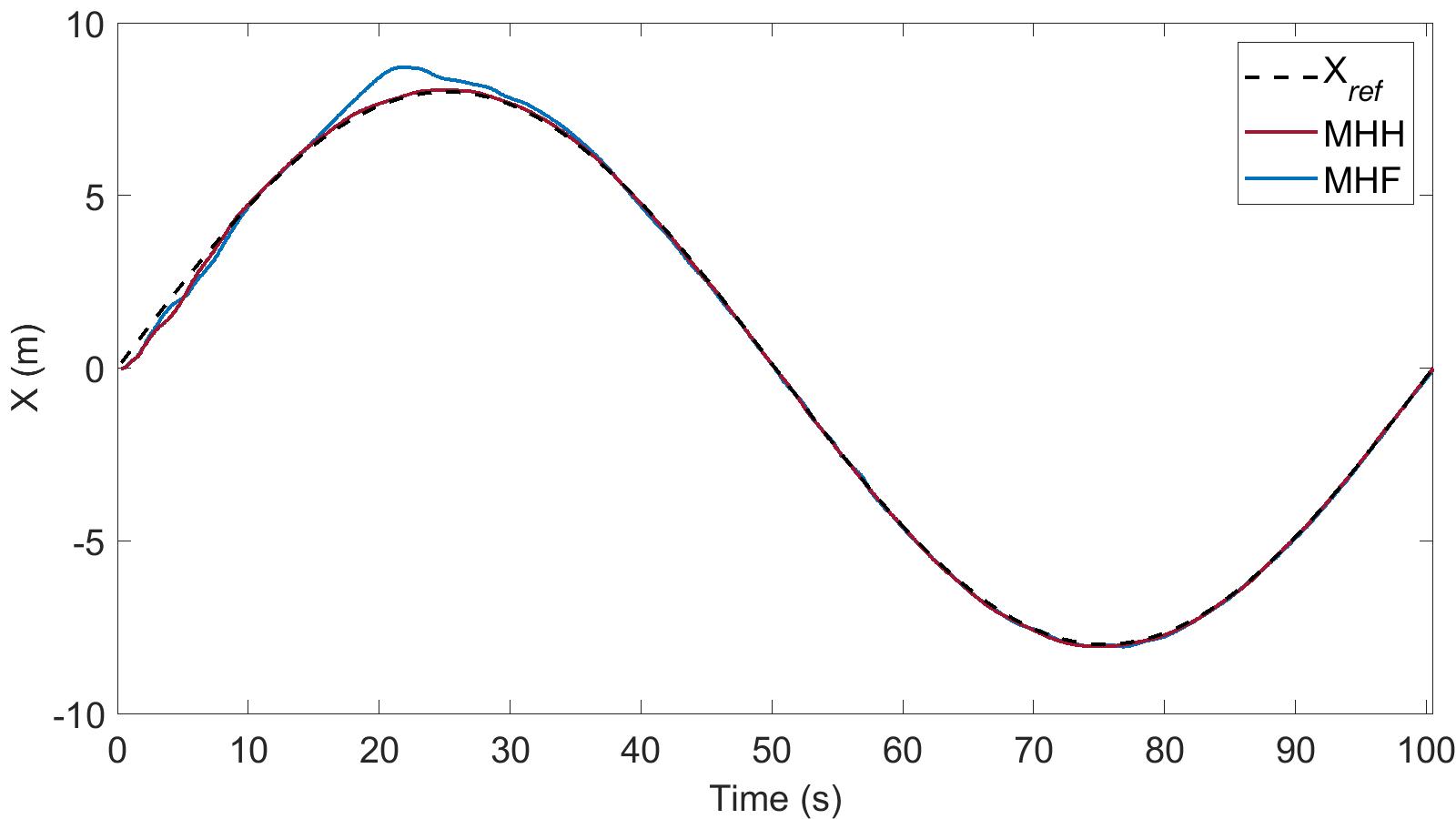}}
    \hspace{0in}
\subfigure[Tracking of $Y$]
{
    \label{fig:subfig:e}
    \includegraphics[height=2.2cm]{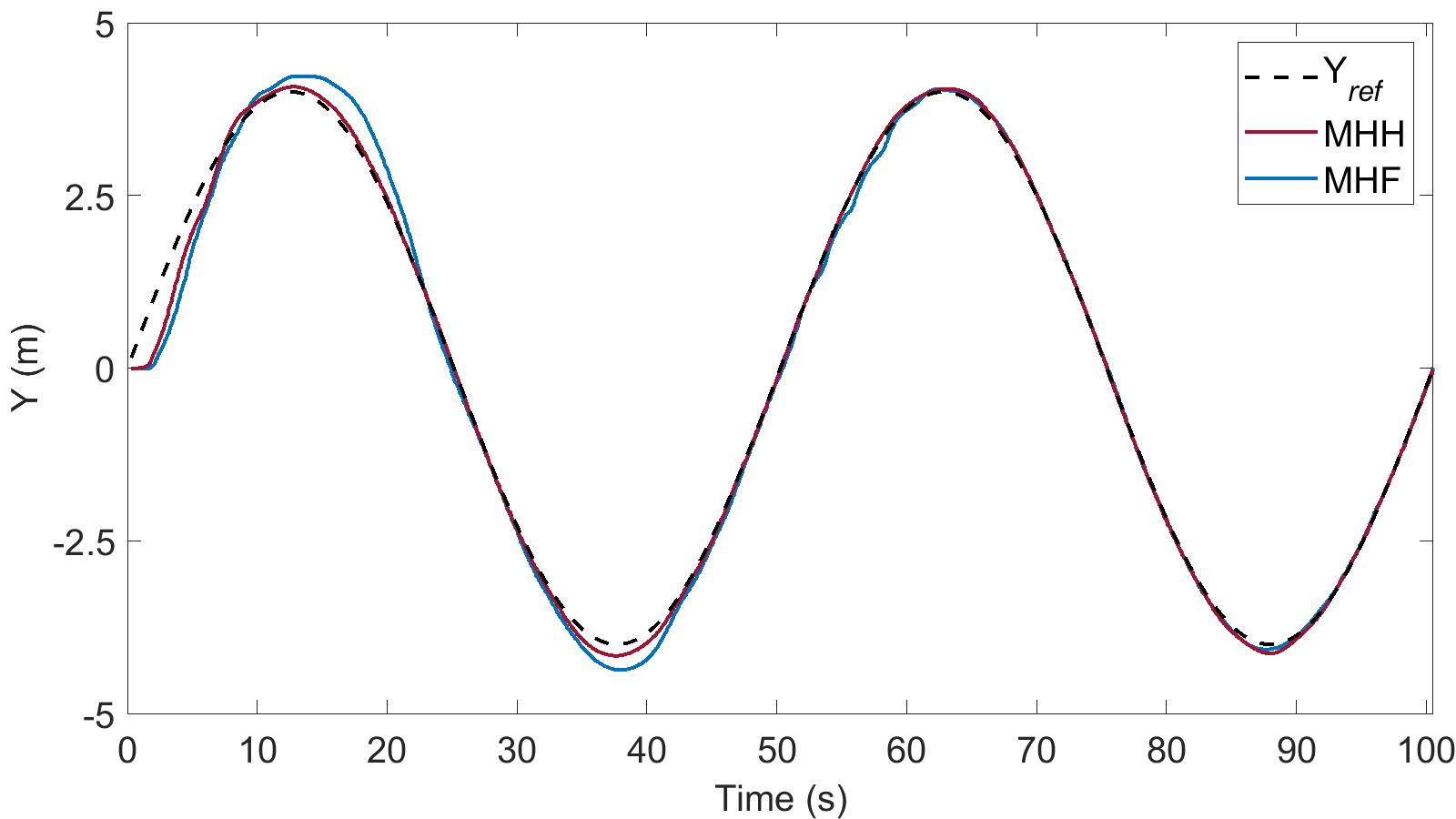}}
    \hspace{0in}
\subfigure[Tracking of $\phi$]
{
    \label{fig:subfig:f}
    \includegraphics[height=2.2cm]{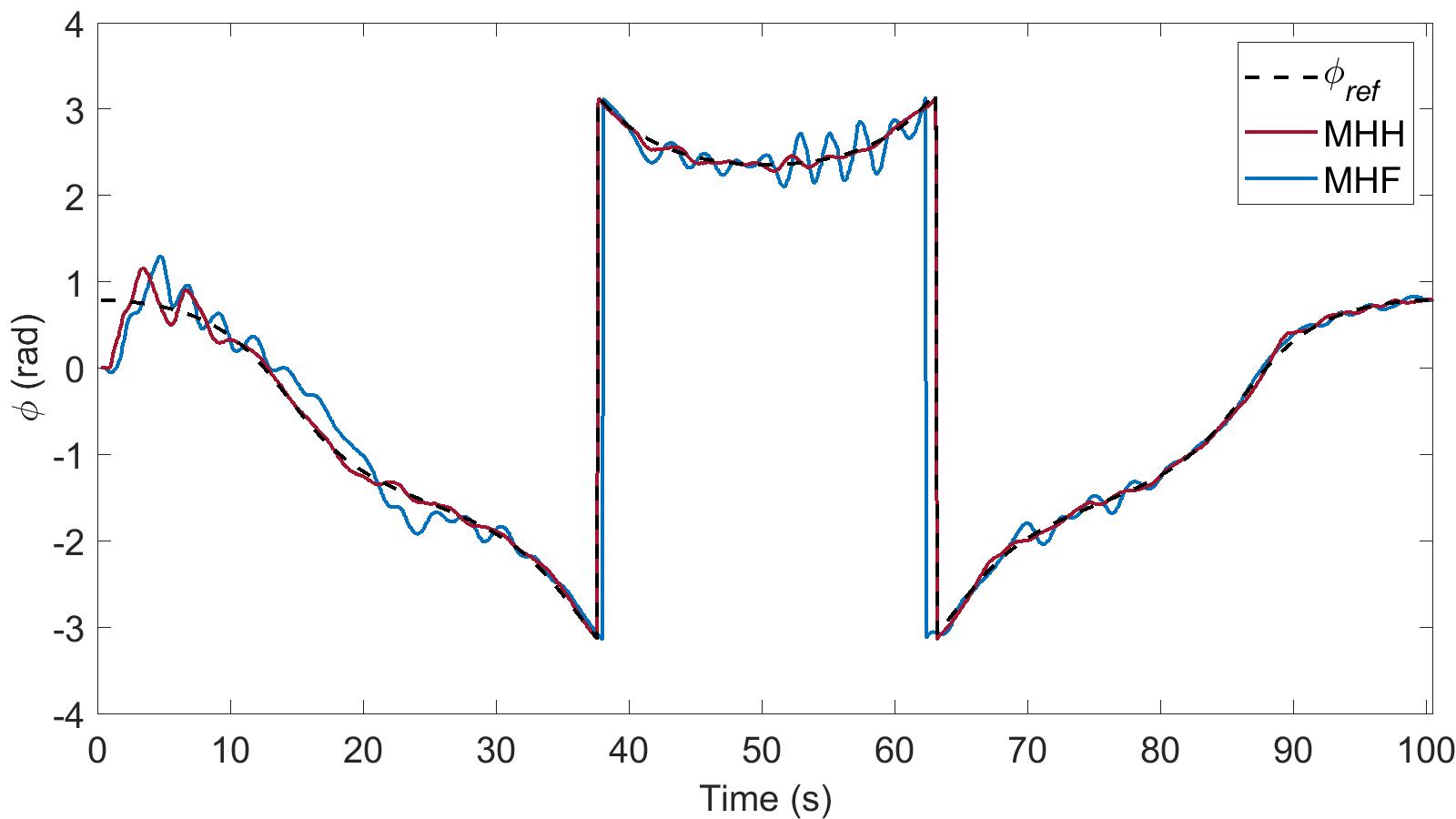}}
\caption{Tracking the trajectory of a  lemniscate of gerono.}
\label{Traj_3}
\vspace{-0.5cm}
\end{figure*}

\subsubsection{Sine-wave Tracking of Roll Angle Controller}
The expected time-variant roll angle in this part is a sine-wave like function which is also shown in \eqref{eq_ex1}. We also apply velocity controller HSMC to keep the robot's velocity at 0.5m/s. Final results are shown in Fig.~\ref{fig_ex3}.
\begin{equation}
q_{rd} = 
\begin{cases}
0&,t<0.1\text{s}\\
10\sin{\left(0.15t-0.015\right)}&, t>=0.1\text{s}
\end{cases}
\label{eq_ex1}
\end{equation}

To further compare the control performance of the two controllers, we again select their root mean square error value $e_{rmse}$ and mean absolute error $e_{mae}$ of the tracking roll angle during the entire procedure to represent the tracking effect. The outcomes of the two indicators are likewise displayed in Table. ~\ref{table3}.

Based on the results in Fig.~\ref{fig_ex3} and Table. ~\ref{table3}, HTSCM can track the sine-wave expected angles very well with much smaller $e_{mae}$ and has more stability with smaller $e_{rmse}$. It seems that there is a tracking delay for Fuzzy-PID, which may be owing to its slower response speed and need for time to accumulate. This results also prove that HTSMC can handle continuous roll angle tracking problem. And with such a direction controller, the system can be more stable in trajectory tracking.

\subsection{Trajectory Tracking Experiments}
In this part, three types of trajectories will be tracked to compare two trajectory tracking frameworks. A same instruction planing controller MPC proposed in \uppercase\expandafter{\romannumeral3}.B and a same velocity controller HSMC proposed in \cite{pre1, pre2} are employed for these two tracking frameworks. And the main difference is the direction controller, with MHH employing HTSMC and MHF Fuzzy-PID. Experiments in this subsection attempt to elucidate the effect of MPC and MHH on the control of the trajectory.

For all of the trajectories, the initial state $\boldsymbol{\overline{X}}=\begin{bmatrix} X & Y & \phi \end{bmatrix}^T = \begin{bmatrix} 0 & 0 & 0 \end{bmatrix}^T$  and $\boldsymbol{\overline{U}}=\begin{bmatrix} v & q_r \end{bmatrix}^T = \begin{bmatrix}0 & 0\end{bmatrix}^T$. Parameters of the MPC are set as below. $N=10$, $\boldsymbol{Q}=diag\begin{bmatrix} 2 & 2 & 1 \end{bmatrix}$, $\boldsymbol{R} = diag \begin{bmatrix}1 & 1\end{bmatrix}$. Green arrays in figures below represent the initial state. Trajectories can all be transformed into math formulation as follows, and experiment results are shown as below as well. Furthermore, CasADi framework, SQPmethod and qpOASEs are employed to solve the nonlinear program in \eqref{eq20}.

\subsubsection{Tracking a Line}
\begin{equation}
\begin{cases}
{X_{ref}}&=0.5t\\
{Y_{ref}}&=2\\
{\phi_{ref}}&=0
\label{tra_ex1}
\end{cases}
\end{equation}
where $t\in[0, 20]$. Results are shown in Fig.~\ref{Traj_1}.

\subsubsection{Tracking a Sine-wave}
\begin{equation}
\begin{cases}
{X_{ref}}&=0.5t\\
{Y_{ref}}&=2\sin{(0.25t)}\\
{\phi_{ref}} &= \arctan{\cos(0.25t)}
\label{tra_ex3}
\end{cases}
\end{equation}
where $t\in[0, 12\pi]$. Results are shown in Fig.~\ref{Traj_2}.

\subsubsection{Tracking a Lemniscate of Gerono}
\begin{equation}
\begin{cases}
{X_{ref}}&=8\sin{(t/16)}\\
{Y_{ref}}&=8\sin{(t/16)}\cos{(t/16)}\\
{\phi_{ref}} &= \text{atan2}\left(\cos{(t/8)}/\cos{(t/16)} S_{ign}, \;\;  S_{ign}\right)\\
S_{ign} &= \text{sgn}\left(\cos{(t/16)}\right)
\label{Traj_3}
\end{cases}
\end{equation}
where $t\in[0, 32\pi]$, $\text{atan2}(\cdot)$ is a function in C++ which is a special $\arctan(\cdot)$ function with returns in $[-\pi, \pi]$. Results are shown in Fig.~\ref{Traj_3}.

\subsubsection{Analysis}
According to Fig.~\ref{Traj_1}, Fig.~\ref{Traj_2} and Fig.~\ref{Traj_3}, it is obvious that MHH can track trajectories considerably better than MHF with less errors, especially when the robot is turning continuously. MHH's true curve is also smoother and has less fluctuation. According to distance figures, it can be prove that MHH can track the expected trajectory faster. Based on the figures of $\phi$ tracking, MHH appears to be more stable than MHF, with less oscillation and a smoother slope.

From the experiments, it also can be imply that HTSMC has better control effect in direction tracking compared with Fuzzy-PID. HTSMC is more stable, maybe because its design is based on the Lyapunov function. As a result, stability constraints are not required for MPC to remain stable. Overall, MHH is an effective trajectory tracking framework for spherical robots that are non-holonomic, non-linear, and under-actuate.

\section{CONCLUSIONS}
A new direction controller, HTSMC, is devised in this work to enable spherical robots follow expected roll angles more stable and quickly. Hierarchical algorithm, quick terminal sliding mode controller, and spherical robot motion features make up this controller. Furthermore, we present an instruction planning controller MPC, and ultimately, we combine two sliding mode controllers and a model prediction controller to create a spherical robot's trajectory tracking framework MHH.

Then we carry out a series of experiments on hardware. The new direction controller provides greater control performance and stability, according to the results. The new trajectory tracking framework is more stable thanks to two Lyapunov-based sliding mode controllers, which eliminate the need for external stability restrictions for MPC. Finally, with the new controller and the trajectory, the spherical robot can have broader applications.

\addtolength{\textheight}{-12cm}   %

\bibliographystyle{IEEEtran}
\bibliography{refer}

\end{document}